\documentclass{article}

\PassOptionsToPackage{numbers}{natbib}
\usepackage[preprint]{neurips_2025}

\usepackage[utf8]{inputenc} % allow utf-8 input
\usepackage[T1]{fontenc}    % use 8-bit T1 fonts
\usepackage{hyperref}       % hyperlinks
\usepackage{url}            % simple URL typesetting
\usepackage{booktabs}       % professional-quality tables
\usepackage{amsfonts}       % blackboard math symbols
\usepackage{nicefrac}       % compact symbols for 1/2, etc.
\usepackage{microtype}      % microtypography
\usepackage{xcolor}         % colors
\usepackage{graphicx}
\usepackage[nolist]{acronym}
\usepackage{subcaption}
\usepackage{tikz}
\usepackage{amsmath}
\usepackage{amssymb}
\usepackage{multirow}
\usetikzlibrary{positioning}
\usepackage[capitalize,noabbrev]{cleveref}
\usepackage{wrapfig}
\usepackage{comment} % remove later

\title{Locality-Sensitive Hashing for Efficient Hard Negative Sampling in Contrastive Learning}

\author{%
  Fabian Deuser\thanks{Institute of Distributed Intelligent Systems, University of the Bundeswehr Munich, Neubiberg, Germany} \\
  \texttt{fabian.deuser@unibw.de} \\
  \And
  Philipp Hausenblas\thanks{Professorship of Data Analytics \& Statistics, University of the Bundeswehr Munich, Neubiberg, Germany} \\
  \texttt{philipp.hausenblas@unibw.de} \\
  \And
  Hannah Schieber\thanks{Technical University of Munich, Human-Centered Computing and Extended Reality Lab, TUM University Hospital, Clinic for Orthopedics and Sports Orthopedics, and Munich Institute of Robotics and Machine Intelligence (MIRMI), Munich, Germany} \\
  \texttt{hannah.schieber@tum.de} \\
  \AND
  Daniel Roth\footnotemark[3] \\
  \texttt{daniel.roth@tum.de} \\
  \And
  Martin Werner\thanks{Professorship of Big Geospatial Data Management, Technical University of Munich, Munich, Germany} \\
  \texttt{martin.werner@tum.de} \\
  \And
  Norbert Oswald\footnotemark[1] \\
  \texttt{norbert.oswald@unibw.de} \\
}

\begin{document}
\begin{acronym}[Bspwwww.]  % Längstes Kürzel in der nachfolgenden
                       % Liste um die Breite der Spalte für die
                       % Abkürzungen zu bestimmen.

%% Eintrag: \acro{Referenzname}[Kürzel]{Langform}
%% Im Text wird die Abkürzung dann mit \ac{Referenzname} benutzt.
%A
\acro{ann}[ANN]{Approximated Nearest Neighbor}
%B
\acro{clip}[CLIP]{Contrastive Language-Image Pre-Training}
%C
%D
%E
%F false negatives
\acro{fn}[FN]{False Negative}
%G
%H
\acro{hn}[HN]{Hard Negative}
\acro{hns}[HNS]{Hard Negative Sampling}
%I
%J
%K
%L
\acro{lsh}[LSH]{Locality-Sensitive Hashing}
%M
%N
\acro{nn}[NN]{Nearest Neighbor}
%Natural Language Processing (NLP)
\acro{nlp}[NLP]{Natural Language Processing}
\acro{pca}[PCA]{Principal Component Analysis}
\end{acronym}

\maketitle

\begin{abstract}
Contrastive learning is a representational learning paradigm in which a neural network maps data elements to feature vectors. It improves the feature space by forming lots with an anchor and examples that are either positive or negative based on class similarity. Hard negative examples, which are close to the anchor in the feature space but from a different class, improve learning performance. Finding such examples of high quality efficiently in large, high-dimensional datasets is computationally challenging. In this paper, we propose a GPU-friendly Locality-Sensitive Hashing (LSH) scheme that quantizes real-valued feature vectors into binary representations for approximate nearest neighbor search. We investigate its theoretical properties and evaluate it on several datasets from textual and visual domain. Our approach achieves comparable or better performance while requiring significantly less computation than existing hard negative mining strategies.
\end{abstract}

\section{Introduction}
\label{sec:intro}

Contrastive learning builds on the principle of distinguishing positive (similar) examples from negative (dissimilar) examples, and aims to learn a representation space where similar data points are closer together. Unlike supervised classification, which relies on rigid label boundaries, contrastive learning provides a more flexible strategy for tasks where such boundaries are inappropriate. It has shown success in diverse domains, including person or player re-identification~\cite{hermans2017defense, zhang2020multi, habel2022clipreident}, face verification~\cite{schroff2015facenet}, cross-view geo-localization~\cite{deuser2023sample4geo, zhu2022transgeo, deuser2024optimizing, deuser2023orientation}, sentence and text retrieval~\cite{reimers-2019-sentence-bert, reimers-2020-multilingual-sentence-bert}, multi-modal retrieval~\cite{radford2021learning, zhai2023sigmoid}, and product search~\cite{patel2022recall, an2023unicom}. In the latter case, for example, items like a sweater and a sweatshirt may look nearly identical but belong to different categories. Similarly, in text retrieval, sentences with different phrasing can convey the same meaning. Such cases highlight the need for embeddings that capture fine-grained similarities beyond discrete class labels.

Batch composition strategies are critical in contrastive learning because they directly affect the effectiveness of training~\cite{wu2017sampling}. \ac{hn} - dissimilar examples that are close to the anchor in the embedding space - can significantly improve learning~\cite{wu2017sampling, galanopoulos2021hard, wang2019multi, yuan2017hard, hermans2017defense, cakir2019learningtorank, xuan2020improved}. A common and efficient strategy is within-batch selection, where \acp{hn} are dynamically identified based on predefined criteria during training~\cite{wang2019multi, yuan2017hard}. However, this limits the view to a single batch. In contrast, global \ac{hn} sampling, which selects negatives from the entire dataset, provides a broader view and has been shown to improve performance~\cite{deuser2023sample4geo}. However, with datasets scaling to millions~\cite{radford2021learning} or even billions~\cite{jia2021scaling} of samples, this approach introduces significant computational overhead. Pre-extracted negatives do not adapt to evolving embeddings during training, further limiting their effectiveness. This highlights the need for more efficient global negative mining strategies that balance dynamic adaptability with scalability.

To mitigate the computational cost of pre-epoch \ac{hn} sampling, we propose a lightweight \ac{ann} method based on \acf{lsh}~\cite{charikar2002simrounding}. By encoding approximate embeddings in a compact binary space, our method enables efficient \ac{hn} retrieval with reduced search time and memory cost. We evaluated our approach on six datasets across two modalities and analyzed the hardness and relevance of the retrieved negatives based on their cosine similarity to true \acp{hn}.

To summarize, we contribute:

\begin{itemize}
    \item A lightweight and efficient framework for global \ac{hn} sampling using \ac{lsh}, enabling scalable training with dynamic embeddings.
    \item A theoretical analysis establishing probabilistic bounds on the similarity of \ac{lsh}-retrieved negatives, and connecting these bounds to observed behavior during training. 
    \item A comprehensive empirical evaluation on six datasets from two modalities, demonstrating the practical effectiveness of our approach in supervised contrastive learning.
\end{itemize}

\section{Related Work}

\subsection{With-In Batch Sampling}

Simo-Serra et al. refined within-batch sampling by selecting \acp{hn} based on loss values computed after the forward step~\cite{simonserra2015disclearn}. Samples are chosen randomly at the start of each epoch, with backward gradients computed only for high-loss cases. \citet{schroff2015facenet} employed the triplet loss in an online mining scheme, selecting \acp{hn} within a batch using $\ell_2$ distance.

Subsequent work introduces semi-\ac{hn} sampling, as mining only the hardest examples can cause model collapse~\cite{wu2017sampling}. Others compared multiple mining strategies for triplet loss in person re-identification, showing that selecting the hardest positive and negative within a batch outperforms prior work~\cite{hermans2017defense, oh2016deep, ding2015deep}. \citet{hermans2017defense} investigated offline hard mining as well. However, selecting the hardest samples across the entire dataset led to suboptimal performance, causing model collapse with standard triplet loss and hindering training. \citet{yuan2017hard} proposed a cascaded model to identify \acp{hn} at different network stages, enabling the model to focus on hard examples when they are the most difficult to distinguish, improving learning effectiveness.

Another strategy is mining informative pairs by comparing negative pairs with the hardest positive pairs and vice versa~\cite{wang2019multi}. \citet{wang2019multi} refined this mining strategy with a soft weighting scheme to more accurately prioritize the selected pairs. For positive sampling \citet{xuan2020improved} found that easiest samples can provide higher generalization as the embedding maintains intra-class variance.

\subsection{Pre-Epoch Sampling}

While previous work focuses on in-batch strategies, \citet{cakir2019learningtorank} take a different approach by defining the batch composition during the training epoch. They use WordNet similarities between classes to determine which classes should be sampled together, effectively introducing harder-to-differentiate samples~\cite{pedersen2004wordnet}. In a previously mentioned study, \citet{hermans2017defense} explored also offline \ac{hn} mining. However, they found \acp{hn} can lead to model collapse with standard triplet loss. 

While most of the previous work focused on image retrieval \citet{gillick2019learning} introduced \ac{hn} sampling for entity retrieval in \ac{nlp}. They encode all mentions and entities, identifying the 10 most similar entities after each epoch. If an incorrect entity ranks higher than the correct one, it is treated as an \ac{hn}. They also used approximate nearest neighbor search, but only to speed up inference. \citet{qu2020rocketqa} enhance this by adding a cross-encoder denoising mechanism to reduce false negatives. Another modality-specific approach was proposed by \citet{karpukhin2020dense} who used a BM25 retriever to search for \acp{hn} within the training corpus.

\citet{xiong2020approximate} pioneered the use of \ac{ann} by storing embeddings in a database during training and performing sampling on asynchronously updated indices. However, their method resulted in a doubling of the GPU usage and thus a significant increase in the computational overhead. In cross-view geo-localization, \citet{deuser2023sample4geo} highlight the significant performance improvements achievable with \ac{hn} sampling, and show that the InfoNCE loss avoids the collapsing problems often associated with triplet loss. However, their method incurs significant computational and storage costs because of the need to compute the entire similarity matrix.

\section{Method}
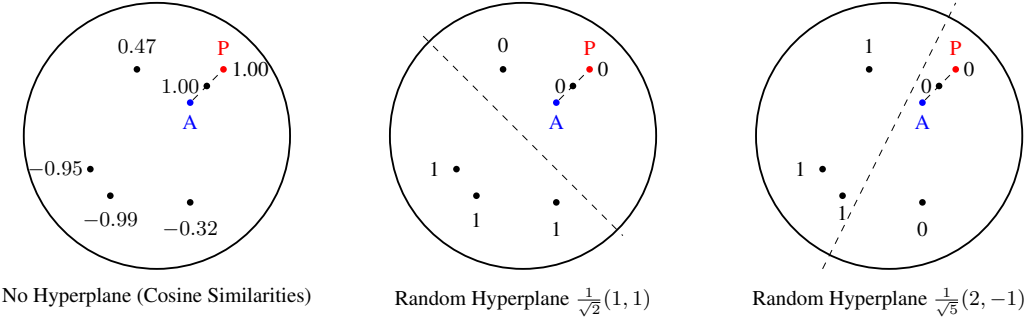
\begin{figure*}[t!]
    \centering
    \resizebox{\textwidth}{!}{
        \begin{tikzpicture}[scale=1.0]
    
    %-----------------------------------------------------------------------
    % 1) LEFT SPHERE: No hyperplane, show Cosine similarities
    %-----------------------------------------------------------------------
            \begin{scope}[xshift=-5.5cm]
              \draw[thick] (0,0) circle(2cm);
    
              % Anchor (A)
              \node[circle,fill=blue,inner sep=1pt] (A) at (0.5,0.5) {};
              % We attach a label slightly below/left:
              \node[below =0pt of A, text=blue] {\small A};
    
              % Positive (P)
              \node[circle,fill=red,inner sep=1pt] (P) at (1,1) {};
              \node[above =1pt of P, text=red] {\small P};
    
              % N1..N5 with numeric labels
              \node[circle,fill=black,inner sep=1pt] (N1) at (0.75,0.75) {};
              \node[left=-2pt of N1] {\small $1.00$};
    
              \node[circle,fill=black,inner sep=1pt] (N2) at (-1,-0.5) {};
              \node[left=-2pt of N2] {\small $-0.95$};
    
              \node[circle,fill=black,inner sep=1pt] (N3) at (-0.7,-0.9) {};
              \node[below=2pt of N3] {\small $-0.99$};
    
              \node[circle,fill=black,inner sep=1pt] (N4) at (0.5,-1) {};
              \node[below=3pt of N4] {\small $-0.32$};
    
              \node[circle,fill=black,inner sep=1pt] (N5) at (-0.3,1) {};
              \node[above=2pt of N5] {\small $0.47$};
    
              % Dashed line from A to P, label "Cos=1.00"
              \draw[dashed] (A) -- (P)
                node[ right=0pt] {\small 1.00};
    
              % Caption
              \node[below,yshift=-1ex] at (0,-2) 
                {\small No Hyperplane (Cosine Similarities)};
            \end{scope}
    
    %-----------------------------------------------------------------------
    % 2) MIDDLE SPHERE: hyperplane y=-x => sign(x+y)
    %    Show 1-bit Hamming dist from A
    %-----------------------------------------------------------------------
            \begin{scope}
              \draw[thick] (0,0) circle(2cm);
    
              % Hyperplane y=-x
              \draw[dashed] (-1.5,1.5) -- (1.5,-1.5);
    
              % Points:
              \node[circle,fill=blue,inner sep=1pt] (A) at (0.5,0.5) {};
              \node[below =0pt of A, text=blue] {\small A};
    
              \node[circle,fill=red,inner sep=1pt] (P) at (1,1) {};
              \node[above =1pt of P, text=red] {\small P};
    
              \node[circle,fill=black,inner sep=1pt] (N1) at (0.75,0.75) {};
              \node[left=-2pt of N1] {\small 0};
    
              \node[circle,fill=black,inner sep=1pt] (N2) at (-1,-0.5) {};
              \node[left=2pt of N2] {\small 1};
    
              \node[circle,fill=black,inner sep=1pt] (N3) at (-0.7,-0.9) {};
              \node[below=2pt of N3] {\small 1};
    
              \node[circle,fill=black,inner sep=1pt] (N4) at (0.5,-1) {};
              \node[below=3pt of N4] {\small 1};
    
              \node[circle,fill=black,inner sep=1pt] (N5) at (-0.3,1) {};
              \node[above=2pt of N5] {\small 0};
    
              \draw[dashed] (A) -- (P)
                node[ right=0pt] {\small 0};
    
              \node[below,yshift=-1ex] at (0,-2) 
                {\small Random Hyperplane $\frac{1}{\sqrt{2}}(1, 1)$};
            \end{scope}
    
    %-----------------------------------------------------------------------
    % 3) RIGHT SPHERE: hyperplane y=2x => sign(2x - y)
    %    Show 1-bit Hamming dist from A
    %-----------------------------------------------------------------------
            \begin{scope}[xshift=5.5cm]
              \draw[thick] (0,0) circle(2cm);
    
              % Hyperplane y=2x
              \draw[dashed] (-1,-2) -- (1,2);
    
              % Points:
              \node[circle,fill=blue,inner sep=1pt] (A) at (0.5,0.5) {};
              \node[below =0pt of A, text=blue] {\small A};
    
              \node[circle,fill=red,inner sep=1pt] (P) at (1,1) {};
              \node[above =1pt of P, text=red] {\small P};
    
              \node[circle,fill=black,inner sep=1pt] (N1) at (0.75,0.75) {};
              \node[left=-2pt of N1] {\small 0};
    
              \node[circle,fill=black,inner sep=1pt] (N2) at (-1,-0.5) {};
              \node[left=2pt of N2] {\small 1};
    
              \node[circle,fill=black,inner sep=1pt] (N3) at (-0.7,-0.9) {};
              \node[below=-1pt of N3] {\small 1};
    
              \node[circle,fill=black,inner sep=1pt] (N4) at (0.5,-1) {};
              \node[below=3pt of N4] {\small 0};
    
              \node[circle,fill=black,inner sep=1pt] (N5) at (-0.3,1) {};
              \node[above=2pt of N5] {\small 1};
    
              \draw[dashed] (A) -- (P)
                node[ right=0pt] {\small 0};
    
              \node[below,yshift=-1ex] at (0,-2) 
                {\small Random Hyperplane $\frac{1}{\sqrt{5}}(2,-1)$};
            \end{scope}
    
        \end{tikzpicture}
    }
    \caption{Illustration of the anchor (A, blue), positive (P, red), and several negatives (N, black). 
    \textbf{Left:} The raw cosine similarities between the anchor and negatives are shown, commonly used to identify \acp{hn}. % (examples very close to the anchor).
    \textbf{Middle and Right:} Two randomly sampled hyperplanes, $\frac{1}{\sqrt{2}}(1, 1)$ and $\frac{1}{\sqrt{5}}(2,-1)$, illustrate that \acp{hn} are likely to be mapped to the same side of the hyperplane as the anchor.
    The Hamming distance, defined by the number of hyperplanes separating embeddings, decreases with higher cosine similarity, enabling effective \ac{hn} identification.}
    \label{fig:hyperplanes}
\end{figure*}

\subsection{Preliminary}

A core challenge in contrastive learning is selecting meaningful positive and negative samples to ensure that positive pairs are embedded closer together than negative pairs. As noted by \cite{schroff2015facenet}, fast convergence depends on ensuring that the similarity $\operatorname{sim}$  between anchor $c$ and positive sample $y_+$ is lower than between anchor and negative samples $y_i$: 

\begin{equation}\operatorname{sim}\left(c, y_+ \right) < \operatorname{sim}\left(c, y_i \right) \forall i\end{equation}

in our case $\operatorname{sim}$ is the cosine similarity. Given an anchor $c$, we call \begin{equation} y_{-} = \arg\max_{y \in Y:y \neq y_+} \operatorname{sim}( c, y) \end{equation}

its \ac{hn} sample. It is the most similar instance of another class to the anchor. 

Since every element in the dataset can be an anchor with its corresponding batch of \acp{hn}, traditional pre-epoch sampling calculates the full similarity matrix $S$. Each entry $S_{ij} = \operatorname{sim}(y_i, y_j)$ represents the cosine similarity between the pairs of embeddings. The calculation of this matrix is computational expensive and memory-intensive due to its size which scales quadratically to dataset size $M$.

To mitigate these computationally expensive and memory-intensive drawbacks, we employ an \ac{ann} method resulting in reduced computational complexity and time. 
%%%%% END NEW PRELIMINARY

\subsection{Locality Sensitive Hashing}
\label{sec:lsh}

During training, it is essential to query stored vectors after each epoch to identify \acp{hn}. However, storing high-dimensional embeddings for large datasets can be space-intensive and, in distributed environments, can impose significant network overhead. In addition, retrieving and processing these embeddings is computationally expensive. To address this, we adopt a binarization strategy based on \ac{lsh}, which drastically reduces memory consumption while allowing efficient retrieval of \acp{ann}. Although dimensionality reduction techniques such as \ac{pca} are also common, they tend to be outperformed by \ac{lsh} as the number of bits increases~\cite{gong2013itq}, a trend we confirm experimentally in our appendix (\Cref{sec:pcacomp}). In addition, \ac{pca} introduces additional computational overhead because it must be fit to the full training set and requires the storage of real-valued feature projections. These factors conflict with the space efficiency and simplicity that our \ac{lsh}-based approach is designed to achieve. Therefore, we avoid PCA and similar methods. Following previous work, we implement this approach by sampling a random rotation (i.e. the vectors are othonormal) matrix~\cite{sariel2012approx, datar2004pstabledist, andoni2015practical} \(R \in \mathbb{R}^{b \times d}\) where \( d \) denotes the dimensionality of the embedded feature vector $y \in Y$, and $ b $ specifies the bit dimension of the encoded feature vector. The embedded dataset $Y$ is first transformed using the random matrix $R$ and then in every dimension centered around its mean.

\begin{align}
    V = RY - \overline{RY}
\end{align}
We then convert every vector $v\in  V$ into a signed vector representation $\hat{v}$:
\begin{align}
    \hat{v}_i = \operatorname{sign}(v_i), \text{ where } \operatorname{sign}(v_i ) \left\{\begin{matrix}
        1 \text{ if } v_i \geq 0\\
        -1 \text{ if } v_i < 0
    \end{matrix}\right.
\end{align} 

Following the work from Wang et al.~\cite{wang2015learning} the probability of an anchor point $c$ and another point $y$ to be mapped into the same bit in one dimension is:
\begin{align}
\operatorname{Pr}\left[h_i\left(c\right)=h_i\left(y\right)\right]=1-\frac{\theta_{c y}}{\pi}=1-\frac{1}{\pi} \cos ^{-1} \frac{c^{\top} y}{\left\|c\right\|\left\|y\right\|}
\end{align}

where $h_i$ is the function that converts the vector $c$ and $y$ into the $i$-th bit $\hat{v}_i$ of the binary representation as described above and $\theta$ is the angle between $c$ and $y$. It depends on the \(i\)-th row in \(R\).
This is illustrated in \cref{fig:hyperplanes} where we show how the cosine similarity between the anchor and the data points affects different hyperplanes $h_i$.
The Hamming distance between $c$ and $y$ \begin{align}
\operatorname{HammDist}(c,y) = \sum_{i=1}^b  \mathbf{1}_{h_i(c)\neq h_i(y)}\end{align}
is the number of bits they differ in. We then use the \(K\) data points with smallest hamming distance to the anchor point \(c\) as \ac{hn}. In our \Cref{sec:lshdesignchoices} we investigate the impact of the used transformations.

Encoding embeddings as binary vectors offers substantial memory and computational benefits for large datasets. Converting from $d$-dimensional 32-bit floats (4 bytes per value) to 1-bit binary representations reduces storage by a factor of 32. This not only lowers memory usage for \ac{hn} sampling, but also enables fast similarity search using Hamming distance, which relies on efficient bitwise operations (XOR and popcount)~\cite{wang2015learning}. These are hardware-accelerated and significantly faster than cosine similarity, particularly in high-dimensional spaces.

\subsubsection{Locality Sensitive Hashing Analysis}
\label{sec:lshtheoanalyis}

We provide a compact theoretical analysis of \ac{lsh} sampling combined with Hamming distance minimization. This technique was introduced by \citet{wang2015learning} but has not been analyzed in depth. Most other approximate-nearest-neighbor methods based on locality-sensitive hashing, such as \citet{har2012approximate}, use multiple hash tables and identify a point \(y'\) as the nearest neighbor of a query \(c\) if they collide in at least one table. These methods are typically analyzed using an approximation factor \(a\) to ensure that the returned point \(y'\) satisfies

\begin{align}
\|y' - c\|_2 \le a \cdot \min_y \|y - c\|_2
\end{align}

with constant probability. In our experiments, selecting the point that minimizes the Hamming distance often yields closer neighbors than selecting any point that collides in at least one table. However, this approach has remained without a theoretical foundation, an issue we aim to address.

Let \(c, x, y\) be points on the unit sphere, \(b\) the bit dimension,
and \(i\in[b]\). In the following steps we want to show that if \(x\) is closer to \(c\) than \(y\) it is likely that also its hamming distance to \(c\) is smaller. 
Let \(X_i, Y_i\) be random variables that take the value \(1\) if \(x\),
respectively \(y\), lie on the opposite side of a randomly drawn
hyperplane from \(c\), and \(0\) otherwise. We remember that the probability of a random hyperplane separating \(c\) and \(x\) is \(\frac{\theta_{cx}}{\pi}\). Then \(X_i\) and \(Y_i\) are Bernoulli distributed with

\begin{equation}
\begin{aligned}
\mu_{X_i} &= \frac{\theta_{cx}}{\pi}, &
\sigma^2_{X_i} &= \bigl(1-\frac{\theta_{cx}}{\pi}\bigr)\,\frac{\theta_{cx}}{\pi},\\[4pt]
\mu_{Y_i} &= \frac{\theta_{cy}}{\pi}, &
\sigma^2_{Y_i} &= \bigl(1-\frac{\theta_{cy}}{\pi}\bigr)\,\frac{\theta_{cy}}{\pi}.
\end{aligned}
\end{equation}

Let \(Z_i = X_i - Y_i\) and
\(\operatorname{Pr}(X_i=1, Y_i=1)\) be the probability that the randomly drawn hyperplane puts both
\(x\) and \(y\) on the opposite side of \(c\).
Then
\begin{equation}
\begin{aligned}
\mu_{Z_i} &= \mu_{X_i} - \mu_{Y_i}
          = \frac{\theta_{cx}}{\pi}-\frac{\theta_{cy}}{\pi}  \\
\sigma^2_{Z_i} &= \sigma^2_{X_i} + \sigma^2_{Y_i}
               - 2\operatorname{Cov}(X_i,Y_i)  \\
\operatorname{Cov}(X_i,Y_i) &=
  \operatorname{Pr}(X_i=1, Y_i=1)
  - \frac{\theta_{cx}}{\pi}
   \frac{\theta_{cy}}{\pi}
\end{aligned}\end{equation}

Set \(Z=\sum_{i=1}^b Z_i\), then  
\(Z = \operatorname{HammDist}(c,x) - \operatorname{HammDist}(c,y)\).
Because the \(Z_i\) are i.i.d. (the hyperplanes are drawn independently), the central limit theorem gives, for
large \(d\), with \(\Phi\) denoting the cumulative distribution function of the standard normal distribution

\begin{equation}
\begin{aligned}
P(Z \le 0) \;\approx\;
\Phi\!\left(
  \frac{- \mu_Z}{\sigma_Z}
\right),
\end{aligned}\end{equation}
with
\begin{equation}
\begin{aligned}
\mu_Z &= b\bigl(\frac{\theta_{cx}}{\pi}-\frac{\theta_{cy}}{\pi}\bigr),
\\
\sigma_Z^2 &= b\!\left(
  \bigl(1-\frac{\theta_{cx}}{\pi}\bigr)\frac{\theta_{cx}}{\pi} +
  \bigl(1-\frac{\theta_{cy}}{\pi}\bigr)\frac{\theta_{cy}}{\pi}
   - 2\!\bigl[\operatorname{Pr}(X_i=1, Y_i=1)-\frac{\theta_{cx}}{\pi}  \frac{\theta_{cy}}{\pi}\bigr] \right).
\end{aligned}
\end{equation}

Hence, we can evaluate the probability that
\(\operatorname{HammDist}(c,x) \le \operatorname{HammDist}(c,y)\)
by normalizing their difference.

Now, assume
\(
\frac{\theta_{cx}}{\pi}=\varepsilon,
\frac{\theta_{cy}}{\pi}=a\varepsilon,
\)
for some small \(\varepsilon\in(0,1)\) and \(a> 1\). That means \(y\) is further away from \(c\) than \(x\) by the factor \(a\).
Then using \(\operatorname{Pr}(X_i=1, Y_i=1)\le \operatorname{Pr}(X_i=1)= \varepsilon\), we get:

\begin{equation}
\begin{aligned}
P(Z\le 0)
  &\ge \Phi\!\left(
      \frac{(a-1)b\varepsilon}
           {\sqrt{\,b\bigl((1-\varepsilon)\varepsilon +
                          (1-a\varepsilon)a\varepsilon -
                          2(\varepsilon - a\varepsilon^2)\bigr)}}
    \right)\\
 % &= \Phi\!\left(
 %     \frac{(a-1)b\varepsilon}
 %          {\sqrt{\,b\varepsilon
 %                  \bigl(a+1 - (a^2+1)\varepsilon -
 %                        2(1-a\varepsilon)\bigr)}}
 %   \right)\\
 % &= \Phi\!\left(
 %     \frac{(a-1)b\varepsilon}
 %          {\sqrt{\,b\varepsilon
 %                  \bigl(a-1 - (a-1)^2\varepsilon\bigr)}}
 %   \right)\\
  &= \Phi\!\left(
      \frac{\sqrt{(a-1)b\varepsilon}}
           {\sqrt{\,1-(a-1)\varepsilon}}
    \right)
    \;\ge\;
    \Phi\!\bigl(\sqrt{(a-1)b\varepsilon}\bigr).
\end{aligned}\end{equation}

The last expression is monotone growing in \(a\). Therefore, \(P(Z\le 0)\ge\Phi\!\bigl(\sqrt{(a-1)b\varepsilon}\bigr)\) holds for \(\frac{\theta_{cy}}{\pi}\ge  a\varepsilon\) too. As this bound is currently defined for three points, namely $x,y,c$ we further give an intuition how to generalize this bound for multiple points $y_i$.
Given the anchor \(c\) and \(\varepsilon\in(0,1)\), let \(x\) be such that \(\frac{\theta_{cy}}{\pi}=\varepsilon\),
and suppose there are \(n\) points \(y_i\) with
\(\frac{\theta_{cy_i}}{\pi}\ge a\varepsilon\). Then we have \(P(\operatorname{HammDist}(c,x)<\operatorname{HammDist}(c,y_i))\ge \Phi\!\bigl(\sqrt{(a-1)b\varepsilon}\bigr)\) for all \(i\in [n]\).
Given some constant \(f>1\) we want to find \(b_{\min}\) such that

\begin{equation}
\Phi\!\bigl(\sqrt{(a-1)\,b_{\min}\,\varepsilon}\bigr) \;\ge\; 1-\frac{1}{fn}.
\end{equation}
Let \(z_n = \Phi^{-1}(1-\frac{1}{fn})\); then

\begin{equation}
b_{\min}
  = \Bigl\lceil
      \frac{z_n^{\,2}}{(a-1)\varepsilon}
    \Bigr\rceil.
\end{equation}
Since \(z_n \approx \sqrt{2\ln fn}\) as \(n \to \infty\),

\begin{equation}
b_{\min}(n)
  \approx \frac{2\ln fn}{(a-1)\varepsilon},
  \qquad n\to\infty.
\end{equation}

Thus the probability that \(\operatorname{HammDist}(c,x)
> \operatorname{HammDist}(c,y_i)\) for \(y_i\) in that farther tier is at most \(\frac{1}{fn}\). Using the union bound, the probability of at least one \(y_i\) having a smaller Hamming Distance to \(c\) than \(x\) is bounded by the constant \(\frac{1}{f}\). We are interested to find for a given \(K\) (our batch size) approximate nearest neighbors of \(c\). Now, we assume that for all nearest neighbors \(x_j, j\in[b]\) it holds that \(\frac{\theta_{cx_j}}{\pi} = \varepsilon_j\le\varepsilon\). Let \(a>1\), then for every \(\varepsilon\) there exists \(a_j\ge a\) such that \(\varepsilon_ja_j=a\varepsilon\). Because \(\)

\begin{equation}
\begin{aligned}
    \Phi(\sqrt{(a_j-1)b_{\min}\varepsilon_j})\ge\Phi(\sqrt{(a-1)b_{\min}\varepsilon}),
\end{aligned}
\end{equation}

for \(b_{\min}\) as defined above, we are guaranteed to find for every \(x_j\) with a probability of at least \(1-\frac{1}{f}\) an \(x'_j\) such that \(\theta_{cx'_j}\le a\varepsilon\). The number of approximate neighbors \(x'_j\) with \(\theta_{cyx'_j}\le a\varepsilon\) thus follows a binomial distribution \(B(b_{\min},1-\frac{1}{f})\).

\subsubsection{Empirical Bound Evaluation}
\label{sec:empiricalbound}

\begin{wrapfigure}{r}{0.45\linewidth}
    \centering
    \includegraphics[width=\linewidth]{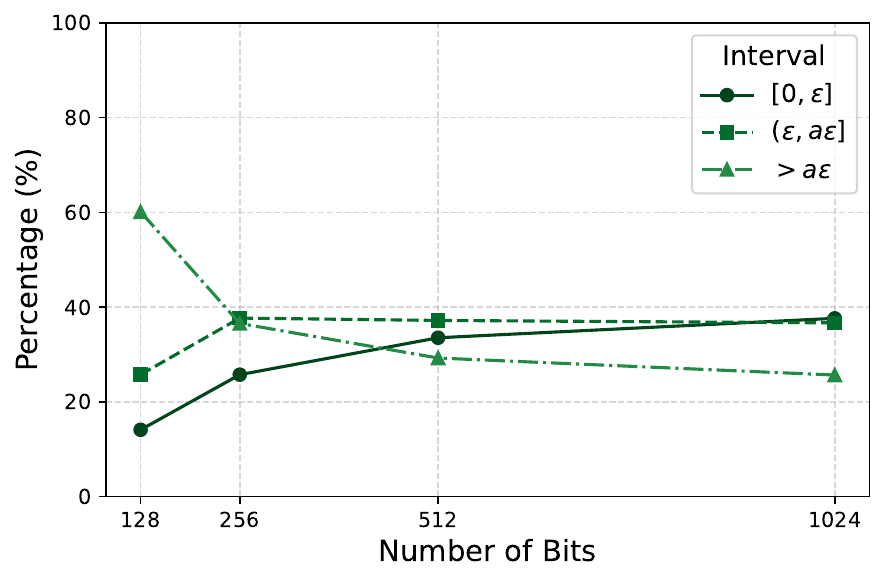}
    \caption{Percentage of neighbors retrieved by \ac{lsh} that fall within the theoretical similarity bounds $\varepsilon$ and $\varepsilon a$, compared to cosine similarity. }
    \label{fig:boundusage}
\end{wrapfigure}

In this experiment, we evaluate the quality of \acp{ann} retrieved using \ac{lsh} for \ac{hn} mining. We use features extracted from the CVUSA dataset~\citet{workmann2015cvusa} using a ConvNeXt-based model pre-trained on ImageNet, without any contrastive fine-tuning. We define a similarity threshold~$\varepsilon$ as the cosine similarity of the 128\textsuperscript{th}-ranked \acp{hn}, corresponding to the batch size. Based on this threshold, we analyze how many of the \ac{lsh} retrieved neighbors fall into three intervals: $[0, \varepsilon]$, $(\varepsilon, a\varepsilon]$, and $> a\varepsilon$, where $a = 1.1$. As shown in the plot, the proportion of retrieved samples within $[0, \varepsilon]$ increases with the number of bits, indicating improved alignment with true \acp{hn}. At the same time, the proportion of distant neighbors ($> a\varepsilon$) decreases. Notably, there remains a gap between the neighbors identified by exact cosine similarity and those found by \ac{lsh}. While neighbors outside the $\varepsilon$ boundary may be less similar, they remain within a controlled range ($a\varepsilon$) and can still contribute meaningfully to the contrastive learning process. Ultimately, whether \ac{lsh}-based \acp{hn} are sufficient for downstream learning is an empirical question that we investigate in our evaluation.

\section{Evaluation}

% We evaluate our approach on different benchmarks and metrics including evaluating the overlap and mean positional distance while assessing \ac{lsh} against pre-epoch and random sampling.

\subsection{Implementation Details}

We employ a Siamese CNN~\cite{chopra2005simiaritymetric} (ConvNeXt-base~\cite{liu2022convnet}, embedding size 1024) to encode image embeddings and a Transformer (Distill-RoBERTa-base~\cite{Sanh2019DistilBERTAD} embedding size 768) for text embeddings. Training uses the InfoNCE loss~\cite{oord2018representation} with a learnable temperature $\tau$, a weight decay of 0.01, and label smoothing of 0.1. We apply cosine learning rate schedules: 1e-3 for the CNN and 1e-4 for the Transformer over 10 epochs.

To reduce overhead, we reuse training embeddings for pre-epoch sampling, which, despite lagging behind the current model, proves efficient with minimal performance impact. High-dimensional embeddings are projected into lower-dimensional binary space using \ac{lsh}, with bit sizes $b \in {128, 256, 512, 1024}$ for images and $b \in {128, 256, 512, 768}$ for text, defined by a random rotation matrix. After each epoch, a binary index is built, and \acp{hn} are identified via Hamming distance for the next epoch’s batch formation. For datasets without predefined query-reference splits~\cite{oh2016deep,liu2016deepfashion}, the hardest positive per class is selected using Hamming distance.

\subsection{Datasets}

We compare our approach on six datasets, two from the domain of product search~\cite{oh2016deep,liu2016deepfashion}, three from the domain of cross-view geo-localization~\cite{zhu2021vigor,workmann2015cvusa,liu2019lending}, and one textual retrieval dataset~\cite{bajaj2016ms}, as prototypical retrieval tasks. We select them to cover multiple modalities and domains, and to demonstrate scalability, as their sizes range from 30k to 500k samples.

For geo-localization we use \textbf{CVUSA}~\cite{workmann2015cvusa}, \textbf{CVACT}~\cite{liu2019lending} and \textbf{VIGOR} \cite{zhu2021vigor}. As product search datasets we leverage \textbf{Stanford Online Products (SOP)}~\cite{oh2016deep} and \textbf{InShop}~\cite{liu2016deepfashion}, and for text retrieval we leverage \textbf{MS Marco}~\cite{bajaj2016ms}. More details can be found in the appendix. 
\begin{table*}[t!]

   \begin{center}
    \caption{Quantitative comparison between multiple sampling methods on supervised image retrieval dataset. Results are reported for Recall@1 (R@1) and Recall@5 (R@5). \label{tab:recall_comparison}}
    \resizebox{\textwidth}{!}{ 
    \begin{tabular}{l|cc|cc|cc|cc|cc|cc|cc} \hline \hline
       \multirow{2}{*}{Approach} & \multicolumn{2}{c|}{CVUSA} & \multicolumn{2}{c|}{CVACT$_{val}$} & \multicolumn{2}{c|}{CVACT$_{test}$} & \multicolumn{2}{c|}{VIGOR$_{same}$} & \multicolumn{2}{c|}{VIGOR$_{cross}$} & \multicolumn{2}{c|}{SOP} & \multicolumn{2}{c}{InShop}  \\  % \cmidrule(r){2.5-4.5}
        & R@1 & R@5 & R@1 & R@5 & R@1 & R@5 & R@1 & R@5 & R@1 & R@5 & R@1 & R@5 & R@1 & R@5 \\\hline
       Random & 97.68 & 99.63 & 87.46 & 96.46 & 60.17 & 89.35 & 64.58 & 91.19 & 36.06 & 62.96 & 87.55 & 94.70 & 91.93 & 97.92 \\
       BatchHard~\cite{schroff2015facenet} & 97.64 & 99.63 & 87.28 & 96.65 & 60.68 & 89.46 & 66.75 & 92.28 & 36.31 & 63.71 & 87.85 & 94.93 & 91.93 & 97.93 \\
       Pre-Epoch Full & 98.68 & 99.67 & 91.01 & 97.11 & 69.98 & 92.82 & 77.11 & 96.11 & 59.86 & 82.55 & 89.44 & 95.76 & 93.21 & 98.21 \\
       Pre-Epoch Incr. & 98.53 & 99.62 & 90.42 & 97.12 & 68.71 & 92.50 & 76.39 & 96.01 & 57.97 & 81.61 & 89.78 & 95.75 & 93.07 & 98.30 \\ \hline
       \ac{lsh}$_{128}$ (ours) & 98.15 & 99.67 & 89.70 & 96.94 & 66.29 & 91.62 & 74.48 & 95.24 & 53.93 & 79.08 & 89.09 & 95.42 & 92.86 & 98.22 \\
       \ac{lsh}$_{256}$ (ours) & 98.43 & 99.65 & 90.07 & 97.02 & 67.27 & 91.93 & 75.50 & 95.59 & 55.99 & 80.53 & 89.34 & 95.54 & 93.19 & 98.13 \\
       \ac{lsh}$_{512}$ (ours) & 98.54 & 99.68 & 90.45 & 97.15 & 68.20 & 92.29 & 76.35 & 95.76 & 57.22 & 81.25 & 89.60 & 95.65 & 93.11 & 98.12 \\
       \ac{lsh}$_{1024}$ (ours) & 98.60 & 99.65 & 90.82 & 97.24 & 68.75 & 92.60 & 76.51 & 95.88 & 57.69 & 81.53 & 89.64 & 95.78 & 93.31 & 98.22 \\\hline \hline
    \end{tabular}
    }
   \end{center}

\end{table*}
\subsection{Sampling Strategies}

For our evaluation, we compare multiple sampling methods with our proposed \ac{lsh} sampling:

\textbf{Random Sampling} selects a subset of all possible pair permutations, constrained by the number of training epochs. \acp{hn} are only included by chance. The only filtering is at the class level to avoid having multiple instances of the same class within a batch. 

\textbf{Pre-Epoch Full Sampling} precomputes \acp{hn} before each epoch by extracting the full training dataset and selecting negatives based on cosine similarity from the similarity matrix. This sampling is the most resource intensive method, as it requires a reprocessing of the complete dataset.

\begin{wraptable}{r}{0.3\linewidth}
    \begin{center}
    \caption{Quantitative comparison between multiple sampling methods on the MS MARCO dataset. Results are reported for MRR@10.\label{tab:mrr_comparison}
    }
    \resizebox{\linewidth}{!}{
    \begin{tabular}{l|c} \hline \hline
       Approach & MRR@10 \\\hline
       Random & 20.07 \\
       BatchHard & 20.59 \\
       Pre-Epoch Incr. & 26.23 \\ \hline
       LSH$_{128}$ (ours) & 24.41 \\
       LSH$_{256}$ (ours) & 25.67 \\
       LSH$_{512}$ (ours) & 26.42 \\
       LSH$_{768}$ (ours) & 26.44 \\ \hline
       Pre-Epoch Full & 26.24 \\\hline \hline
    \end{tabular}}
    \end{center}
    \vspace{-4em}
\end{wraptable}

\textbf{Pre-Epoch Incremental Sampling}\label{sec:preepoch} extracts \acp{hn} during training using saved embeddings before weight updates. This method is faster compared to \textit{pre-epoch full sampling} but relies on embeddings that might partially not be updated yet.

\textbf{BatchHard Sampling} calculates the loss using only the 50\% hardest negatives within a batch. Since the loss method can influence \ac{hns} selection, we follow Schroff et al.~\cite{schroff2015facenet} and implement BatchHard for the InfoNCE loss. 

\subsection{Impact of Different Sampling Strategies}

%TODO underscore best and second best results? or provide some aggregation at the end like average over all R@1?

In our vision experiments (see \autoref{tab:recall_comparison}), random sampling serves as a baseline and consistently underperforms compared to other strategies, while BatchHard offers modest improvements. Pre-Epoch Full Sampling, while the slowest method, often yields the best performance by extracting embeddings after each epoch. Compared to Pre-Epoch Incremental, \ac{lsh} shows slightly lower performance at low bit dimensions (128/256), but matches its effectiveness at higher dimensions (512/1024), while being faster and more memory efficient since it avoids retaining full embedding vectors.

Similar results are observed for text retrieval, see \autoref{tab:mrr_comparison}: random sampling underperforms, while \ac{hn} improves results. Unlike in vision tasks, higher \ac{lsh} bit counts further enhance performance even beyond Pre-Epoch full sampling.

\subsection{Search Speed Comparison}
\label{subsec:searchspeed}

\begin{wrapfigure}{r}{0.45\linewidth}
    \centering
    \includegraphics[width=\linewidth]{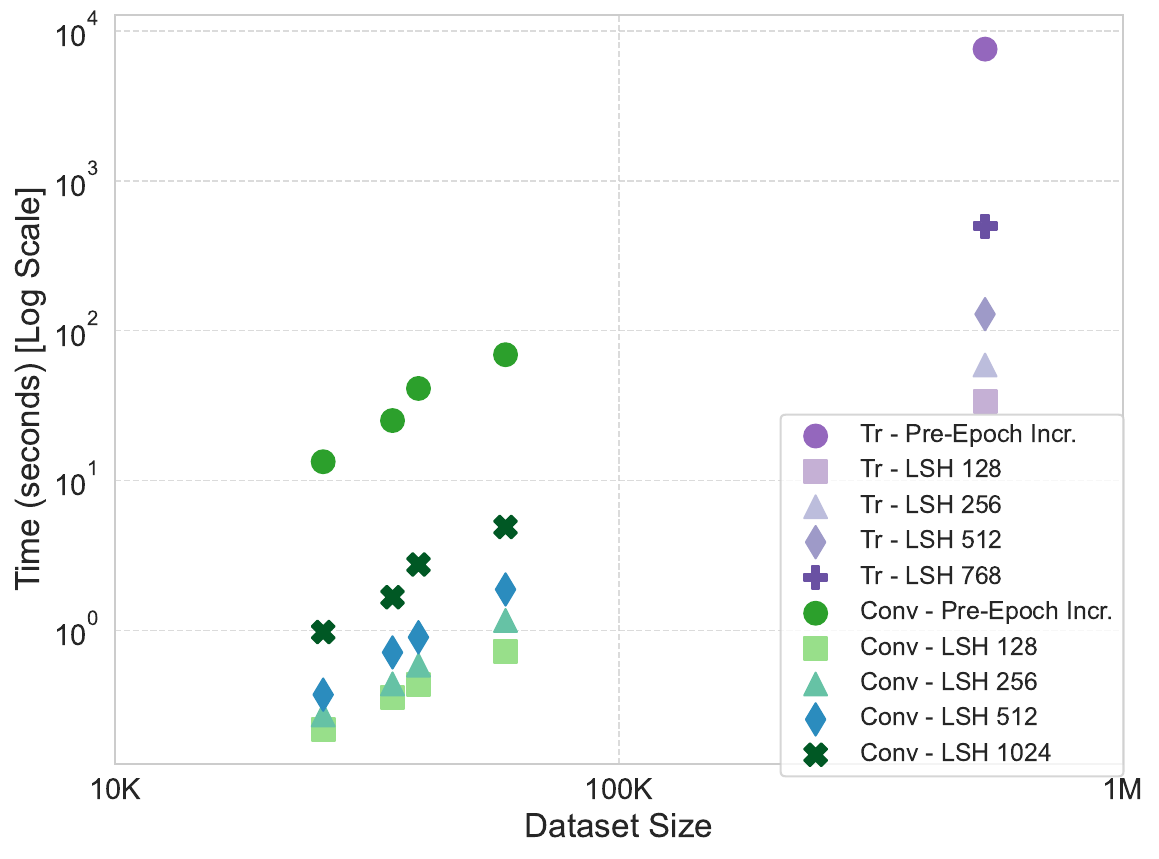}
    \caption{\acs{lsh} and pre-epoch \acs{hn} sampling in comparison considering the search time vs. the dataset size and the model output size.} 
    \label{fig:speed_comparison}
% \end{wrapfigure}

% \begin{wrapfigure}{r}{0.5\linewidth}
%     \centering
    \includegraphics[width=\linewidth]{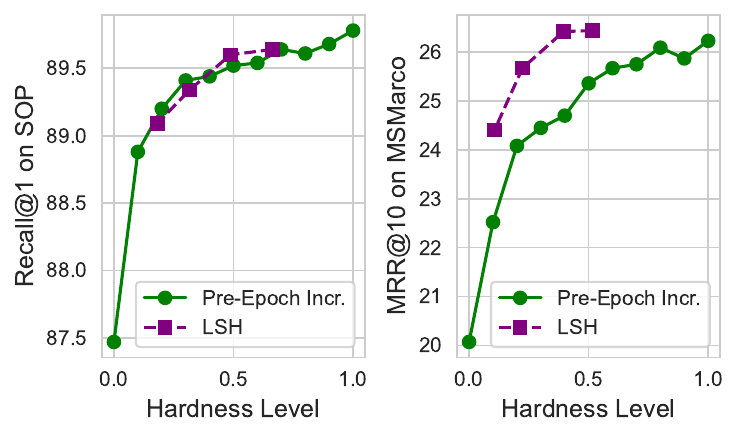}
    \caption{Impact of \ac{hn} hardness on R@1 on SOP (left) and MRR@10 on MSMarco  (right). We define hardness as the percentage of \acp{hn} within a batch and include the results from \ac{lsh} based on the respective overlap.}
    \label{fig:recallvscorruption}
   \vspace{-3em}
\end{wrapfigure}

In \Cref{fig:speed_comparison} we compare the search times of \ac{lsh} (with different bit sizes) against \ac{hn} sampling using full float32 embeddings. Tr denotes the used Transformer and Conv the used ConvNeXt. All searches retrieve the top 128 \acp{nn} using FAISS~\cite{douze2024faiss}. While the theoretical complexity is $O(n \cdot d)$ for all indices, the use of binary features in \ac{lsh} yields substantial speedups. We encode the training data for each dataset, excluding CVACT (identical in size to CVUSA). For datasets with a reference-query split, retrieval is performed within the query set.

On all datasets, including large ones like MS MARCO with over 500K samples, \ac{lsh} consistently outperforms full-precision cosine search in speed. The reported times reflect the per-epoch search overhead during training, where \ac{lsh} dramatically reduces the cost of \ac{hn} mining. In the appendix, we show that \ac{lsh} reduces the additional training time from 9 to over 88 percentage points, with larger savings on larger datasets. 

\subsection{Neighbor Analysis}
\label{sec:nnanalysis}

%We further investigate the behavior of \acp{hn} selection when \ac{lsh} is used. For this analysis, we use the MS-MARCO dataset with over 500k samples and the SOP dataset, which contains over 60k samples, to comprehensively evaluate the generalization of our approach. 
We analyze \acp{hn} selection behavior under \ac{lsh} using the MS-MARCO dataset (500k+ samples) and the SOP dataset (60k+ samples) to evaluate the generalization of our approach. In our appendix we provide results for VIGOR as well. In each setting, we retrieve the top 128 \acp{nn} and rank them based on their similarity to the reference feature. This allows us to compute the mean positional distance between the \acp{ann} obtained by \ac{lsh} and the actual \acp{nn} determined by cosine similarity. Furthermore, we evaluate the overlap between the retrieved neighbors and the real \ac{nn} to quantify the effectiveness of the approximation.

\subsubsection{Neighbor Overlap}
\label{sec:overlap}

\begin{figure}[t!]
    \centering
    \begin{subfigure}[t]{0.5\textwidth}
        \centering
        \includegraphics[width=\textwidth]{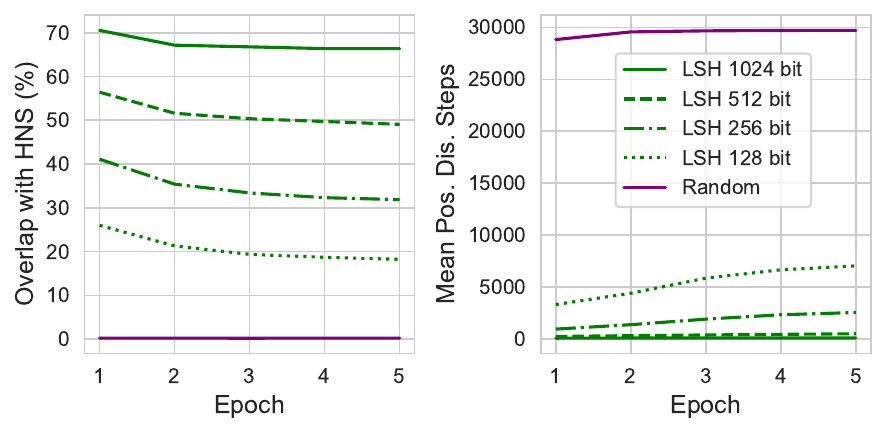}
        \caption{SOP\label{fig:neighbor_overlap_sop}}
    \end{subfigure}%
    ~ 
    \begin{subfigure}[t]{0.5\textwidth}
        \centering
        \includegraphics[width=\textwidth]{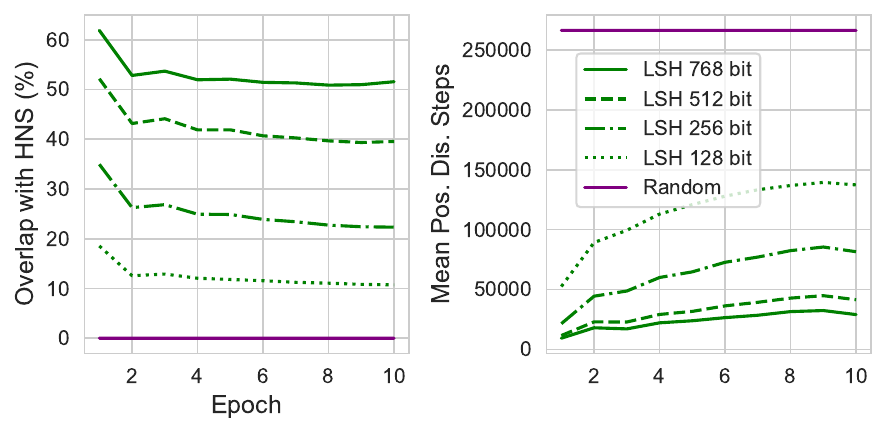}
        \caption{MS Marco\label{fig:overlap_dist_msmarco}}
    \end{subfigure}
    
    \caption{A comparison of \ac{lsh} and random sampling on SOP and MS Marco. We compare the overlap with Pre-Epoch Increment (\acp{hn}) and mean positional distance. 
    } 
    \label{fig:neighbor_overlap}
\end{figure}

For SOP, the overlap with random sampling is approximately one percent, see \Cref{fig:neighbor_overlap}. In contrast, using \ac{lsh} increases the overlap, and our strategy achieves around 70\% overlap at the highest bit count.

Based on the evaluation in \Cref{tab:recall_comparison} and the observed 54\% overlap at 512 bits, this level of precision proves sufficient to achieve strong performance.
At lower bit widths (128 and 256), the overlap decreases over time, a behavior consistent with the loss objective of pushing close negatives apart. This makes true \acp{hn} harder to identify, especially since their selection depends on similarity, see \Cref{sec:lsh}. For the mean positional distance, random sampling converges to the center of the sampled subset of negatives. In contrast, the 1024-bit \ac{lsh} stays near zero, with positional distance increasing as the number of bits decreases. Note that since only a subset of all possible negative permutations is sampled, convergence to the subset's center does not represent a true average of the entire dataset. In theory, random sampling could still select all true \ac{nn}.

The text modality shows a different behavior, as shown in \Cref{fig:overlap_dist_msmarco}. The overlap is much lower than in the vision setting, and the mean positional distance remains high even at larger bit sizes. While true \acp{hn} are still retrievable, the task is more challenging due to the fine-grained and multi-concept nature of text, as opposed to the more coherent semantics of images. Nonetheless, 512-bit \ac{lsh} achieves strong results, outperforming both pre-epoch incremental and full sampling. Additional analysis of cosine similarity between retrieved and true \acp{hn} is provided in the appendix.

\subsubsection{Neighbor Hardness}
\label{sec:neighbor_hardness}

Understanding how many \acp{hn} are needed to maintain good performance is key to evaluating the tradeoffs of using \ac{ann}. As shown in \Cref{sec:nnanalysis}, \ac{lsh} achieves about 70\% overlap with true \acp{hn} at high bit counts. In this experiment, we evaluate how varying the proportion of \acp{hn} within a batch (i.e., hardness) affects performance. The remaining samples in the batch are drawn as random negatives from the data set and are not considered hard. \Cref{fig:recallvscorruption} shows the results after 10 training epochs at different hardness levels, where 1.0 corresponds to incremental sampling before the epoch (all \acp{hn}) and 0.0 corresponds to pure random sampling. We also include \ac{lsh} results based on the overlap from \Cref{fig:neighbor_overlap}. While SOP performance remains stable, \ac{lsh} yields notable gains on MS MARCO, even at equivalent hardness, suggesting that it selects more effective \acp{hn} than random sampling. These results are consistent with our theoretical analysis in \Cref{sec:lshtheoanalyis}. Results for VIGOR are included in the Appendix.

\section{Conclusion}

We show that \ac{lsh} effectively approximates real \acp{nn} and enables robust \ac{hn} mining through locality preservation and similarity-based retrieval. Even at lower bit sizes (256-512), it achieves strong overlap with true \acp{nn} and outperforms traditional methods such as BatchHard~\cite{schroff2015facenet} and random sampling. This demonstrates that \ac{lsh} is a reliable and efficient alternative to \ac{hn} sampling in contrastive learning. Furthermore, we demonstrate that \ac{lsh}-based mining is generalized across different modalities, including vision and text retrieval.

In addition, \ac{lsh} significantly reduces the time and memory overhead compared to pre-epoch sampling, with binarization reducing storage costs and search scaling more favorably for large datasets. Despite these reductions, the quality of \ac{hn} remains competitive or superior. Finally, we explore the theoretical guarantees of \ac{lsh} in the context of \ac{hn} mining and show how these bounds translate into practical performance, bridging theory and practice.

\section{Limitations and Potential Social Impact}
\label{sec:limitations}

Our work focuses on supervised learning, where identifying true \acp{hn} remains a challenge, particularly when positives and \acp{hn} belong to the same class. This overlap can hinder convergence and degrade performance. As seen in \Cref{tab:mrr_comparison}, the pre-epoch full and incremental variants underperform our best \ac{lsh} trainings at higher bit sizes. This may be due to false positives, samples that are semantically similar but incorrectly treated as \acp{hn}. Previous work has explored this issue for unsupervised learning~\cite{robinson2020contrastive, chuang2020debiased}, we leave its resolution in supervised scenarios to future work.

We analyze our \ac{lsh} method theoretically in \Cref{sec:lshtheoanalyis} and demonstrate how a similarity bound can be established and used to retrieve \ac{hn} in \Cref{sec:empiricalbound}. However, the choice of the number of bits is based on empirical evaluation. While it is theoretically possible to estimate the required bit size for a given $\varepsilon$, the actual value is highly dependent on the underlying data distribution and cannot be determined a priori. Therefore, a method that can dynamically adapt or estimate the optimal bit size during training would further strengthen the practicality and robustness of our approach.

Due to ethical concerns around re-identification, we deliberately exclude related models and datasets, and have not trained models for this task. While such methods have applications in areas like sports analytics \cite{habel2022clipreident,zhang2020multi}, they also carry significant risks in surveillance contexts. Our involvement is limited to a literature review. However, our work can contribute positively to the field of contrastive learning by demonstrating a more efficient and scalable approach to \ac{hn} mining, enabling faster and more accessible model training.

\bibliography{main}

\begin{thebibliography}{53}
\providecommand{\natexlab}[1]{#1}
\providecommand{\url}[1]{\texttt{#1}}
\expandafter\ifx\csname urlstyle\endcsname\relax
  \providecommand{\doi}[1]{doi: #1}\else
  \providecommand{\doi}{doi: \begingroup \urlstyle{rm}\Url}\fi

\bibitem[An et~al.(2023)An, Deng, Yang, Li, Feng, Guo, Yang, and Liu]{an2023unicom}
Xiang An, Jiankang Deng, Kaicheng Yang, Jaiwei Li, Ziyong Feng, Jia Guo, Jing Yang, and Tongliang Liu.
\newblock Unicom: Universal and compact representation learning for image retrieval.
\newblock \emph{arXiv preprint arXiv:2304.05884}, 2023.

\bibitem[Andoni et~al.(2015)Andoni, Indyk, Laarhoven, Razenshteyn, and Schmidt]{andoni2015practical}
Alexandr Andoni, Piotr Indyk, Thijs Laarhoven, Ilya Razenshteyn, and Ludwig Schmidt.
\newblock Practical and optimal lsh for angular distance.
\newblock \emph{Advances in neural information processing systems}, 28, 2015.

\bibitem[Bajaj et~al.(2016)Bajaj, Campos, Craswell, Deng, Gao, Liu, Majumder, McNamara, Mitra, Nguyen, et~al.]{bajaj2016ms}
Payal Bajaj, Daniel Campos, Nick Craswell, Li~Deng, Jianfeng Gao, Xiaodong Liu, Rangan Majumder, Andrew McNamara, Bhaskar Mitra, Tri Nguyen, et~al.
\newblock Ms marco: A human generated machine reading comprehension dataset.
\newblock \emph{arXiv preprint arXiv:1611.09268}, 2016.

\bibitem[Cakir et~al.(2019)Cakir, He, Xia, Kulis, and Sclaroff]{cakir2019learningtorank}
Fatih Cakir, Kun He, Xide Xia, Brian Kulis, and Stan Sclaroff.
\newblock Deep metric learning to rank.
\newblock In \emph{2019 IEEE/CVF Conference on Computer Vision and Pattern Recognition (CVPR)}, pages 1861--1870, 2019.
\newblock \doi{10.1109/CVPR.2019.00196}.

\bibitem[Charikar(2002)]{charikar2002simrounding}
Moses~S. Charikar.
\newblock Similarity estimation techniques from rounding algorithms.
\newblock In \emph{Proceedings of the Thiry-Fourth Annual ACM Symposium on Theory of Computing}, STOC '02, page 380–388, New York, NY, USA, 2002. Association for Computing Machinery.
\newblock ISBN 1581134959.
\newblock \doi{10.1145/509907.509965}.
\newblock URL \url{https://doi.org/10.1145/509907.509965}.

\bibitem[Chopra et~al.(2005)Chopra, Hadsell, and LeCun]{chopra2005simiaritymetric}
S.~Chopra, R.~Hadsell, and Y.~LeCun.
\newblock Learning a similarity metric discriminatively, with application to face verification.
\newblock In \emph{2005 IEEE Computer Society Conference on Computer Vision and Pattern Recognition (CVPR'05)}, volume~1, pages 539--546 vol. 1, 2005.
\newblock \doi{10.1109/CVPR.2005.202}.

\bibitem[Chuang et~al.(2020)Chuang, Robinson, Lin, Torralba, and Jegelka]{chuang2020debiased}
Ching-Yao Chuang, Joshua Robinson, Yen-Chen Lin, Antonio Torralba, and Stefanie Jegelka.
\newblock Debiased contrastive learning.
\newblock \emph{Advances in neural information processing systems}, 33:\penalty0 8765--8775, 2020.

\bibitem[Datar et~al.(2004)Datar, Immorlica, Indyk, and Mirrokni]{datar2004pstabledist}
Mayur Datar, Nicole Immorlica, Piotr Indyk, and Vahab~S. Mirrokni.
\newblock Locality-sensitive hashing scheme based on p-stable distributions.
\newblock In \emph{Proceedings of the Twentieth Annual Symposium on Computational Geometry}, SCG '04, page 253–262, New York, NY, USA, 2004. Association for Computing Machinery.
\newblock ISBN 1581138857.
\newblock \doi{10.1145/997817.997857}.
\newblock URL \url{https://doi.org/10.1145/997817.997857}.

\bibitem[Deuser et~al.(2023{\natexlab{a}})Deuser, Habel, and Oswald]{deuser2023sample4geo}
Fabian Deuser, Konrad Habel, and Norbert Oswald.
\newblock Sample4geo: Hard negative sampling for cross-view geo-localisation.
\newblock In \emph{ICCV}, pages 16847--16856, 2023{\natexlab{a}}.

\bibitem[Deuser et~al.(2023{\natexlab{b}})Deuser, Habel, Werner, and Oswald]{deuser2023orientation}
Fabian Deuser, Konrad Habel, Martin Werner, and Norbert Oswald.
\newblock Orientation-guided contrastive learning for uav-view geo-localisation.
\newblock In \emph{Proceedings of the 2023 Workshop on UAVs in Multimedia: Capturing the World from a New Perspective}, pages 7--11, 2023{\natexlab{b}}.

\bibitem[Deuser et~al.(2024)Deuser, Werner, Habel, and Oswald]{deuser2024optimizing}
Fabian Deuser, Martin Werner, Konrad Habel, and Norbert Oswald.
\newblock Optimizing geo-localization with k-means re-ranking in challenging weather conditions.
\newblock In \emph{Proceedings of the 2nd Workshop on UAVs in Multimedia: Capturing the World from a New Perspective}, pages 9--13, 2024.

\bibitem[Ding et~al.(2015)Ding, Lin, Wang, and Chao]{ding2015deep}
Shengyong Ding, Liang Lin, Guangrun Wang, and Hongyang Chao.
\newblock Deep feature learning with relative distance comparison for person re-identification.
\newblock \emph{Pattern Recognition}, 48\penalty0 (10):\penalty0 2993--3003, 2015.

\bibitem[Dosovitskiy(2020)]{dosovitskiy2020image}
Alexey Dosovitskiy.
\newblock An image is worth 16x16 words: Transformers for image recognition at scale.
\newblock \emph{arXiv preprint arXiv:2010.11929}, 2020.

\bibitem[Douze et~al.(2024)Douze, Guzhva, Deng, Johnson, Szilvasy, Mazaré, Lomeli, Hosseini, and Jégou]{douze2024faiss}
Matthijs Douze, Alexandr Guzhva, Chengqi Deng, Jeff Johnson, Gergely Szilvasy, Pierre-Emmanuel Mazaré, Maria Lomeli, Lucas Hosseini, and Hervé Jégou.
\newblock The faiss library.
\newblock 2024.

\bibitem[Fu et~al.(2019)Fu, Xiang, Wang, and Cai]{chong2019hsnw}
Cong Fu, Chao Xiang, Changxu Wang, and Deng Cai.
\newblock Fast approximate nearest neighbor search with the navigating spreading-out graph.
\newblock \emph{Proc. VLDB Endow.}, 12\penalty0 (5):\penalty0 461–474, January 2019.
\newblock ISSN 2150-8097.
\newblock \doi{10.14778/3303753.3303754}.
\newblock URL \url{https://doi.org/10.14778/3303753.3303754}.

\bibitem[Galanopoulos and Mezaris(2021)]{galanopoulos2021hard}
Damianos Galanopoulos and Vasileios Mezaris.
\newblock Hard-negatives or non-negatives? a hard-negative selection strategy for cross-modal retrieval using the improved marginal ranking loss.
\newblock In \emph{CVPR}, pages 2312--2316, 2021.

\bibitem[Gillick et~al.(2019)Gillick, Kulkarni, Lansing, Presta, Baldridge, Ie, and Garcia-Olano]{gillick2019learning}
Daniel Gillick, Sayali Kulkarni, Larry Lansing, Alessandro Presta, Jason Baldridge, Eugene Ie, and Diego Garcia-Olano.
\newblock Learning dense representations for entity retrieval.
\newblock \emph{arXiv preprint arXiv:1909.10506}, 2019.

\bibitem[Gong et~al.(2013)Gong, Lazebnik, Gordo, and Perronnin]{gong2013itq}
Yunchao Gong, Svetlana Lazebnik, Albert Gordo, and Florent Perronnin.
\newblock Iterative quantization: A procrustean approach to learning binary codes for large-scale image retrieval.
\newblock \emph{IEEE Transactions on Pattern Analysis and Machine Intelligence}, 35\penalty0 (12):\penalty0 2916--2929, 2013.
\newblock \doi{10.1109/TPAMI.2012.193}.

\bibitem[Habel et~al.(2022)Habel, Deuser, and Oswald]{habel2022clipreident}
Konrad Habel, Fabian Deuser, and Norbert Oswald.
\newblock Clip-reident: Contrastive training for player re-identification.
\newblock In \emph{Proceedings of the 5th International ACM Workshop on Multimedia Content Analysis in Sports}, MMSports '22, page 129–135, New York, NY, USA, 2022. Association for Computing Machinery.
\newblock ISBN 9781450394888.
\newblock \doi{10.1145/3552437.3555698}.
\newblock URL \url{https://doi.org/10.1145/3552437.3555698}.

\bibitem[Har-Peled et~al.(2012{\natexlab{a}})Har-Peled, Indyk, and Motwani]{har2012approximate}
Sariel Har-Peled, Piotr Indyk, and Rajeev Motwani.
\newblock Approximate nearest neighbor: Towards removing the curse of dimensionality.
\newblock 2012{\natexlab{a}}.

\bibitem[Har-Peled et~al.(2012{\natexlab{b}})Har-Peled, Indyk, and Motwani]{sariel2012approx}
Sariel Har-Peled, Piotr Indyk, and Rajeev Motwani.
\newblock Approximate nearest neighbor: Towards removing the curse of dimensionality.
\newblock \emph{Theory of Computing}, 8\penalty0 (14):\penalty0 321--350, 2012{\natexlab{b}}.
\newblock \doi{10.4086/toc.2012.v008a014}.
\newblock URL \url{https://theoryofcomputing.org/articles/v008a014}.

\bibitem[Hermans et~al.(2017)Hermans, Beyer, and Leibe]{hermans2017defense}
Alexander Hermans, Lucas Beyer, and Bastian Leibe.
\newblock In defense of the triplet loss for person re-identification.
\newblock \emph{arXiv preprint arXiv:1703.07737}, 2017.

\bibitem[Jia et~al.(2021)Jia, Yang, Xia, Chen, Parekh, Pham, Le, Sung, Li, and Duerig]{jia2021scaling}
Chao Jia, Yinfei Yang, Ye~Xia, Yi-Ting Chen, Zarana Parekh, Hieu Pham, Quoc Le, Yun-Hsuan Sung, Zhen Li, and Tom Duerig.
\newblock Scaling up visual and vision-language representation learning with noisy text supervision.
\newblock In \emph{International conference on machine learning}, pages 4904--4916. PMLR, 2021.

\bibitem[Jégou et~al.(2011)Jégou, Douze, and Schmid]{jegou2011pq}
Herve Jégou, Matthijs Douze, and Cordelia Schmid.
\newblock Product quantization for nearest neighbor search.
\newblock \emph{IEEE Transactions on Pattern Analysis and Machine Intelligence}, 33\penalty0 (1):\penalty0 117--128, 2011.
\newblock \doi{10.1109/TPAMI.2010.57}.

\bibitem[Karpukhin et~al.(2020)Karpukhin, Oguz, Min, Lewis, Wu, Edunov, Chen, and Yih]{karpukhin2020dense}
Vladimir Karpukhin, Barlas Oguz, Sewon Min, Patrick~SH Lewis, Ledell Wu, Sergey Edunov, Danqi Chen, and Wen-tau Yih.
\newblock Dense passage retrieval for open-domain question answering.
\newblock In \emph{EMNLP (1)}, pages 6769--6781, 2020.

\bibitem[Liu and Li(2019)]{liu2019lending}
Liu Liu and Hongdong Li.
\newblock Lending orientation to neural networks for cross-view geo-localization.
\newblock In \emph{Proceedings of the IEEE/CVF Conference on Computer Vision and Pattern Recognition}, pages 5624--5633, 2019.

\bibitem[Liu(2019)]{liu2019roberta}
Yinhan Liu.
\newblock Roberta: A robustly optimized bert pretraining approach.
\newblock \emph{arXiv preprint arXiv:1907.11692}, 364, 2019.

\bibitem[Liu et~al.(2022)Liu, Mao, Wu, Feichtenhofer, Darrell, and Xie]{liu2022convnet}
Zhuang Liu, Hanzi Mao, Chao-Yuan Wu, Christoph Feichtenhofer, Trevor Darrell, and Saining Xie.
\newblock A convnet for the 2020s.
\newblock In \emph{Proceedings of the IEEE/CVF conference on computer vision and pattern recognition}, pages 11976--11986, 2022.

\bibitem[Liu et~al.(2016)Liu, Luo, Qiu, Wang, and Tang]{liu2016deepfashion}
Ziwei Liu, Ping Luo, Shi Qiu, Xiaogang Wang, and Xiaoou Tang.
\newblock Deepfashion: Powering robust clothes recognition and retrieval with rich annotations.
\newblock In \emph{Proceedings of the IEEE conference on computer vision and pattern recognition}, pages 1096--1104, 2016.

\bibitem[Oh~Song et~al.(2016)Oh~Song, Xiang, Jegelka, and Savarese]{oh2016deep}
Hyun Oh~Song, Yu~Xiang, Stefanie Jegelka, and Silvio Savarese.
\newblock Deep metric learning via lifted structured feature embedding.
\newblock In \emph{CVPR}, pages 4004--4012, 2016.

\bibitem[Oord et~al.(2018)Oord, Li, and Vinyals]{oord2018representation}
Aaron van~den Oord, Yazhe Li, and Oriol Vinyals.
\newblock Representation learning with contrastive predictive coding.
\newblock \emph{arXiv preprint arXiv:1807.03748}, 2018.

\bibitem[Patel et~al.(2022)Patel, Tolias, and Matas]{patel2022recall}
Yash Patel, Giorgos Tolias, and Ji{\v{r}}{\'\i} Matas.
\newblock Recall@ k surrogate loss with large batches and similarity mixup.
\newblock In \emph{Proceedings of the IEEE/CVF Conference on Computer Vision and Pattern Recognition}, pages 7502--7511, 2022.

\bibitem[Pedersen et~al.(2004)Pedersen, Patwardhan, Michelizzi, et~al.]{pedersen2004wordnet}
Ted Pedersen, Siddharth Patwardhan, Jason Michelizzi, et~al.
\newblock Wordnet:: Similarity-measuring the relatedness of concepts.
\newblock In \emph{AAAI}, volume~4, pages 25--29, 2004.

\bibitem[Qu et~al.(2020)Qu, Ding, Liu, Liu, Ren, Zhao, Dong, Wu, and Wang]{qu2020rocketqa}
Yingqi Qu, Yuchen Ding, Jing Liu, Kai Liu, Ruiyang Ren, Wayne~Xin Zhao, Daxiang Dong, Hua Wu, and Haifeng Wang.
\newblock Rocketqa: An optimized training approach to dense passage retrieval for open-domain question answering.
\newblock \emph{arXiv preprint arXiv:2010.08191}, 2020.

\bibitem[Radford et~al.(2021)Radford, Kim, Hallacy, Ramesh, Goh, Agarwal, Sastry, Askell, Mishkin, Clark, et~al.]{radford2021learning}
Alec Radford, Jong~Wook Kim, Chris Hallacy, Aditya Ramesh, Gabriel Goh, Sandhini Agarwal, Girish Sastry, Amanda Askell, Pamela Mishkin, Jack Clark, et~al.
\newblock Learning transferable visual models from natural language supervision.
\newblock In \emph{International conference on machine learning}, pages 8748--8763. PMLR, 2021.

\bibitem[Reimers and Gurevych(2019)]{reimers-2019-sentence-bert}
Nils Reimers and Iryna Gurevych.
\newblock Sentence-bert: Sentence embeddings using siamese bert-networks.
\newblock In \emph{Proceedings of the 2019 Conference on Empirical Methods in Natural Language Processing}. Association for Computational Linguistics, 11 2019.
\newblock URL \url{https://arxiv.org/abs/1908.10084}.

\bibitem[Reimers and Gurevych(2020)]{reimers-2020-multilingual-sentence-bert}
Nils Reimers and Iryna Gurevych.
\newblock Making monolingual sentence embeddings multilingual using knowledge distillation.
\newblock In \emph{Proceedings of the 2020 Conference on Empirical Methods in Natural Language Processing}. Association for Computational Linguistics, 11 2020.
\newblock URL \url{https://arxiv.org/abs/2004.09813}.

\bibitem[Robinson et~al.(2020)Robinson, Chuang, Sra, and Jegelka]{robinson2020contrastive}
Joshua Robinson, Ching-Yao Chuang, Suvrit Sra, and Stefanie Jegelka.
\newblock Contrastive learning with hard negative samples.
\newblock \emph{arXiv preprint arXiv:2010.04592}, 2020.

\bibitem[Sanh et~al.(2019)Sanh, Debut, Chaumond, and Wolf]{Sanh2019DistilBERTAD}
Victor Sanh, Lysandre Debut, Julien Chaumond, and Thomas Wolf.
\newblock Distilbert, a distilled version of bert: smaller, faster, cheaper and lighter.
\newblock \emph{ArXiv}, abs/1910.01108, 2019.

\bibitem[Schroff et~al.(2015)Schroff, Kalenichenko, and Philbin]{schroff2015facenet}
Florian Schroff, Dmitry Kalenichenko, and James Philbin.
\newblock Facenet: A unified embedding for face recognition and clustering.
\newblock In \emph{Proceedings of the IEEE conference on computer vision and pattern recognition}, pages 815--823, 2015.

\bibitem[Simo-Serra et~al.(2015)Simo-Serra, Trulls, Ferraz, Kokkinos, Fua, and Moreno-Noguer]{simonserra2015disclearn}
Edgar Simo-Serra, Eduard Trulls, Luis Ferraz, Iasonas Kokkinos, Pascal Fua, and Francesc Moreno-Noguer.
\newblock Discriminative learning of deep convolutional feature point descriptors.
\newblock In \emph{2015 IEEE International Conference on Computer Vision (ICCV)}, pages 118--126, 2015.
\newblock \doi{10.1109/ICCV.2015.22}.

\bibitem[Wang et~al.(2015)Wang, Liu, Kumar, and Chang]{wang2015learning}
Jun Wang, Wei Liu, Sanjiv Kumar, and Shih-Fu Chang.
\newblock Learning to hash for indexing big data—a survey.
\newblock \emph{Proceedings of the IEEE}, 104\penalty0 (1):\penalty0 34--57, 2015.

\bibitem[Wang et~al.(2019)Wang, Han, Huang, Dong, and Scott]{wang2019multi}
Xun Wang, Xintong Han, Weilin Huang, Dengke Dong, and Matthew~R Scott.
\newblock Multi-similarity loss with general pair weighting for deep metric learning.
\newblock In \emph{Proceedings of the IEEE/CVF conference on computer vision and pattern recognition}, pages 5022--5030, 2019.

\bibitem[Wightman(2019)]{rw2019timm}
Ross Wightman.
\newblock Pytorch image models.
\newblock \url{https://github.com/rwightman/pytorch-image-models}, 2019.

\bibitem[Workman et~al.(2015)Workman, Souvenir, and Jacobs]{workmann2015cvusa}
Scott Workman, Richard Souvenir, and Nathan Jacobs.
\newblock Wide-area image geolocalization with aerial reference imagery.
\newblock In \emph{2015 IEEE International Conference on Computer Vision (ICCV)}, pages 3961--3969, 2015.
\newblock \doi{10.1109/ICCV.2015.451}.

\bibitem[Wu et~al.(2017)Wu, Manmatha, Smola, and Krahenbuhl]{wu2017sampling}
Chao-Yuan Wu, R~Manmatha, Alexander~J Smola, and Philipp Krahenbuhl.
\newblock Sampling matters in deep embedding learning.
\newblock In \emph{CVPR}, pages 2840--2848, 2017.

\bibitem[Xiong et~al.(2020)Xiong, Xiong, Li, Tang, Liu, Bennett, Ahmed, and Overwijk]{xiong2020approximate}
Lee Xiong, Chenyan Xiong, Ye~Li, Kwok-Fung Tang, Jialin Liu, Paul Bennett, Junaid Ahmed, and Arnold Overwijk.
\newblock Approximate nearest neighbor negative contrastive learning for dense text retrieval.
\newblock \emph{arXiv preprint arXiv:2007.00808}, 2020.

\bibitem[Xuan et~al.(2020)Xuan, Stylianou, and Pless]{xuan2020improved}
Hong Xuan, Abby Stylianou, and Robert Pless.
\newblock Improved embeddings with easy positive triplet mining.
\newblock In \emph{Proceedings of the IEEE/CVF Winter Conference on Applications of Computer Vision}, pages 2474--2482, 2020.

\bibitem[Yuan et~al.(2017)Yuan, Yang, and Zhang]{yuan2017hard}
Yuhui Yuan, Kuiyuan Yang, and Chao Zhang.
\newblock Hard-aware deeply cascaded embedding.
\newblock In \emph{Proceedings of the IEEE international conference on computer vision}, pages 814--823, 2017.

\bibitem[Zhai et~al.(2023)Zhai, Mustafa, Kolesnikov, and Beyer]{zhai2023sigmoid}
Xiaohua Zhai, Basil Mustafa, Alexander Kolesnikov, and Lucas Beyer.
\newblock Sigmoid loss for language image pre-training.
\newblock In \emph{Proceedings of the IEEE/CVF International Conference on Computer Vision}, pages 11975--11986, 2023.

\bibitem[Zhang et~al.(2020)Zhang, Wu, Yang, Wu, Chen, and Xu]{zhang2020multi}
Ruiheng Zhang, Lingxiang Wu, Yukun Yang, Wanneng Wu, Yueqiang Chen, and Min Xu.
\newblock Multi-camera multi-player tracking with deep player identification in sports video.
\newblock \emph{Pattern Recognition}, 102:\penalty0 107260, 2020.

\bibitem[Zhu et~al.(2021)Zhu, Yang, and Chen]{zhu2021vigor}
Sijie Zhu, Taojiannan Yang, and Chen Chen.
\newblock Vigor: Cross-view image geo-localization beyond one-to-one retrieval.
\newblock In \emph{Proceedings of the IEEE/CVF Conference on Computer Vision and Pattern Recognition}, pages 3640--3649, 2021.

\bibitem[Zhu et~al.(2022)Zhu, Shah, and Chen]{zhu2022transgeo}
Sijie Zhu, Mubarak Shah, and Chen Chen.
\newblock Transgeo: Transformer is all you need for cross-view image geo-localization.
\newblock In \emph{Proceedings of the IEEE/CVF Conference on Computer Vision and Pattern Recognition}, pages 1162--1171, 2022.

\end{thebibliography}

\bibliographystyle{plainnat}

%%%%%%%%%%%%%%%%%%%%%%%%%%%%%%%%%%%%%%%%%%%%%%%%%%%%%%%%%%%%
\newpage
\appendix

\section{Appendix}

\subsection{Datasets}
We briefly describe the used datasets in the respective sizes:
\textbf{CVUSA}~\cite{workmann2015cvusa}, contains images from all over the US from different locations and 35,532 pairs for training and 8,884 in the validation set. 

\textbf{CVACT}~\cite{liu2019lending} contains the same amount of data for training and validation, but in the region of Canberra, Australia, and extends further with a test set containing over 92k images. In CVUSA and CVACT, the street view is always centered on the aerial view. 

\textbf{VIGOR} contains 90,618 aerial views and 105,214 street views with arbitrary positions within the aerial view, significantly increasing the challenge of the task~\cite{zhu2021vigor}. These datasets provide two configurations: "cross" and "equal". In the cross setting, training data is derived from two cities, while testing is performed on the other two cities. Conversely, the same setting uses samples from all four city regions for both training and testing. 

\textbf{Stanford Online Products (SOP)}~\cite{oh2016deep} contains $\approx120,000$ images with 22,634 different classes and nearly a 50:50 (training:testing) split. 

\textbf{InShop}~\cite{liu2016deepfashion} consists of over 52k images with 7,982 different clothing types as classes. 

\textbf{MS Marco}~\cite{bajaj2016ms} is a textual retrieval benchmark with 500,000 training queries and 8.8 million passages based on Bing queries.

\subsection{Similarity Distribution Analysis}
\begin{figure*}[h!]
    \centering

    \begin{subfigure}{\textwidth}
        \includegraphics[width=\textwidth]{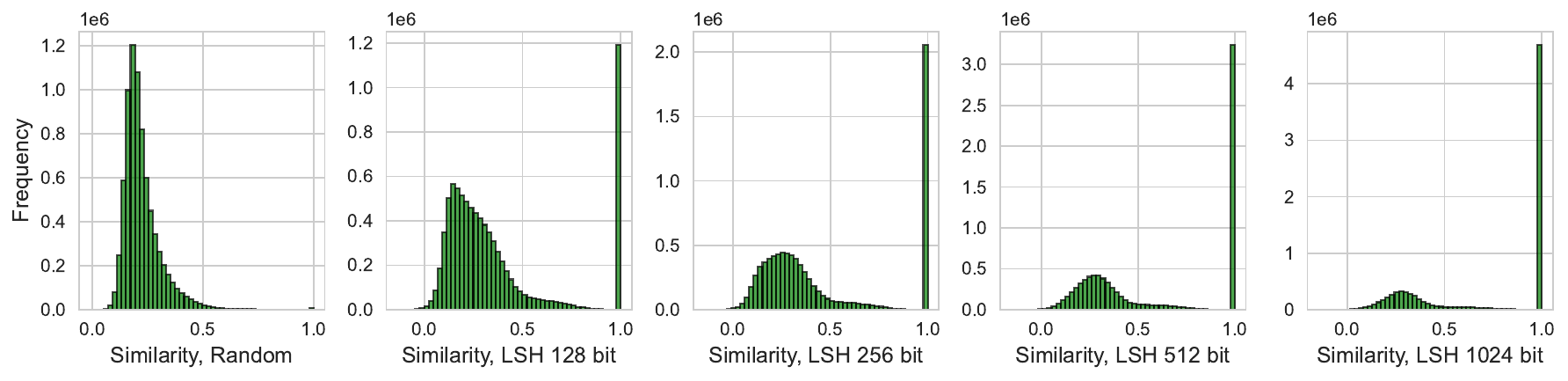}
        \caption{Similarity on SOP across different bit sizes.}
        \label{fig:sim_sop}
    \end{subfigure}
    
    \begin{subfigure}{\textwidth}
        \includegraphics[width=\textwidth]{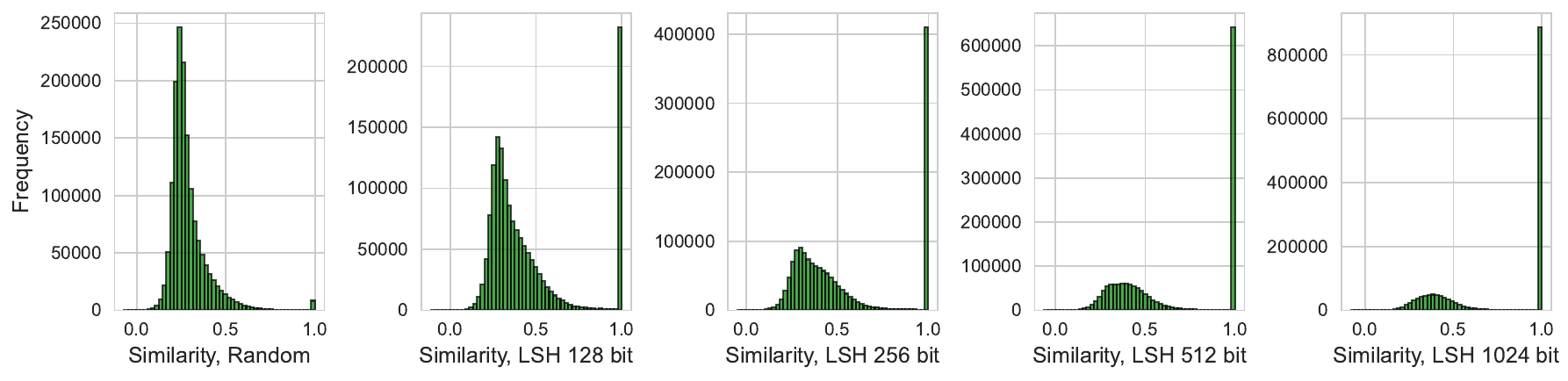}
        \caption{Similarity on VIGOR across different bit sizes.}
        \label{fig:sim_vigor}
    \end{subfigure}
    \begin{subfigure}{\textwidth}
        \includegraphics[width=\textwidth]{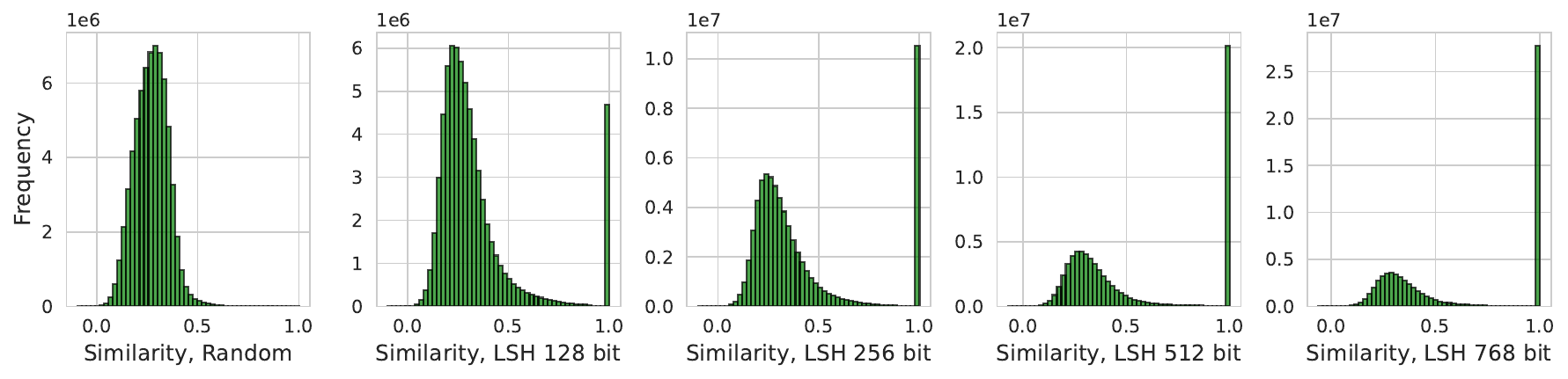}
        \caption{Similarity on MS MARCO across different bit sizes.}
        \label{fig:sim_msmarco}
    \end{subfigure}
    \caption{Comparison of the similarity between the retrieved approximated \acp{hn} and the actual \ac{hn} retrieved by the cosine similarity for SOP (top), VIGOR(middle) and MS MARCO (bottom).}
    \label{fig:similarity_histograms}
\end{figure*}

We further examine the distribution of similarities between the \ac{ann} identified by our sampling method and the real \ac{nn} retrieved by cosine similarity, as shown in \Cref{fig:similarity_histograms}. Notably, while random sampling produces a uniform similarity distribution, increasing the bit count consistently shifts the distribution toward 0.5, regardless of the dataset used. This highlights the advantage of using \ac{lsh}, as the similarity in the original embedding space affects the probability of hash collisions. Even if the true \ac{nn} is not found, the retrieved examples are better than those obtained by random sampling.

\begin{wrapfigure}{r}{0.5\linewidth}
    \centering
    \includegraphics[]{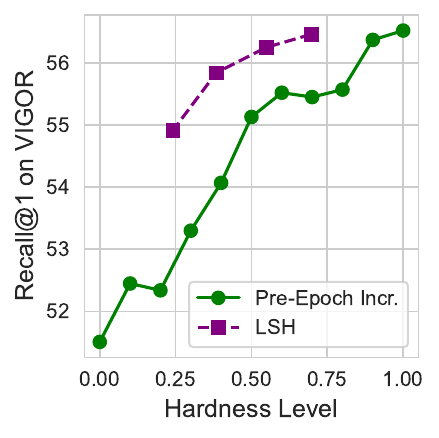}
    \caption{Impact of \ac{hn} hardness on R@1 on VIGOR. We define hardness as the percentage of \acp{hn} within a batch and include the results from LSH based on the respective overlap.}
    \label{fig:hardness_vigor}
    \vspace{-4em}
\end{wrapfigure}
\subsection{VIGOR Analysis}
\begin{figure}[]
    \centering
    \includegraphics[width=0.5\linewidth]{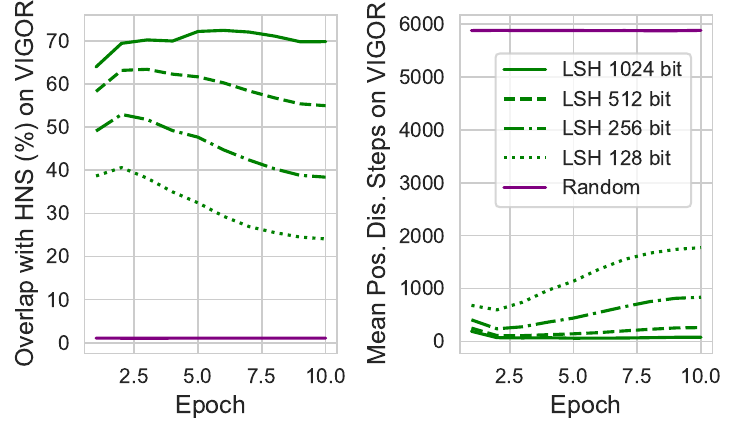}
    \caption{A comparison of overlap and mean positional distance of \ac{lsh} with varying bit sizes (128, 256, 512, and 1024) and random sampling. The overlap with Pre-Epoch Increment (\acp{hn}) and mean positional distance (right) on VIGOR.} % TODO better explain!
    \label{fig:neighbor_overlap_vigor}
\end{figure}

Furthermore, we also compare the overlap and mean positional distance a subset of VIGOR in \cref{fig:neighbor_overlap_vigor}. In this subset we only use the city of Seattle for training and evaluation. Similar to the other datasets the overlap declines over time as the embeddings of pairs are pushed afar from each other. 
We also investigate the impact of hardness during training in \cref{fig:hardness_vigor}, increasing hardness of the sample improves the Recall@1 on the VIGOR dataset. We further include the results achieved with LSH based on our overlap depicted in \Cref{fig:neighbor_overlap_vigor}. Similar to MS MARCO we achieve higher values of Recall@1 while the overlap remains the same.

%%%%%%%%%%%%%%%%%%%%%%%%%%%%%%%%%%%%%%%%%%%%%%%%%%%%%%%%%%%%%%%%%%%%%%%%%%%%%%%
%%%%%%%%%%%%%%%%%%%%%%%%%%%%%%%%%%%%%%%%%%%%%%%%%%%%%%%%%%%%%%%%%%%%%%%%%%%%%%%
\subsection{Further Implementation Details}
\label{sec:impldetails}
For our image-based experiments, we apply several data augmentation techniques during training, including flipping, rotation, coarse dropout, grid dropout, and color jitter. These augmentations help improve model generalization by introducing variability into the training samples. 

In supervised settings, where multiple positive pairs exist for a given label, we structure our batches to contain only one positive pair per label. This approach minimizes redundancy and reduces noise in the loss computation, ensuring a more stable training process.

We train for 40 epochs on the CVUSA, CVACT, and VIGOR datasets, which focus on cross-view geo-localization tasks. For datasets with a different retrieval structure, such as Stanford Online Products (SOP), InShop, and the MS MARCO text retrieval dataset, we limit training to 10 epochs to avoid overfitting. For the cross-view dataset, we resize the images for CVUSA and CVACT to $384 \times 384$ for the satellite image and $112 \times 616$ for the street view image, for VIGOR we use $384 \times 384$ for the satellite view and $384 \times 768$ for the street view. In the experiments for SOP and InShop, all images are resized to $384 \times 384$. 

For all our experiments we use a DGX-2 with 16 Nvidia V100 graphics cards. For all experiments, 4 GPUs are used, except for training with VIGOR, where the image size is larger and we needed 8 GPUs to handle the large batch sizes in memory.
\subsection{Training Process:} 
\begin{figure}[th!]
    \centering
    \includegraphics[width=\linewidth]{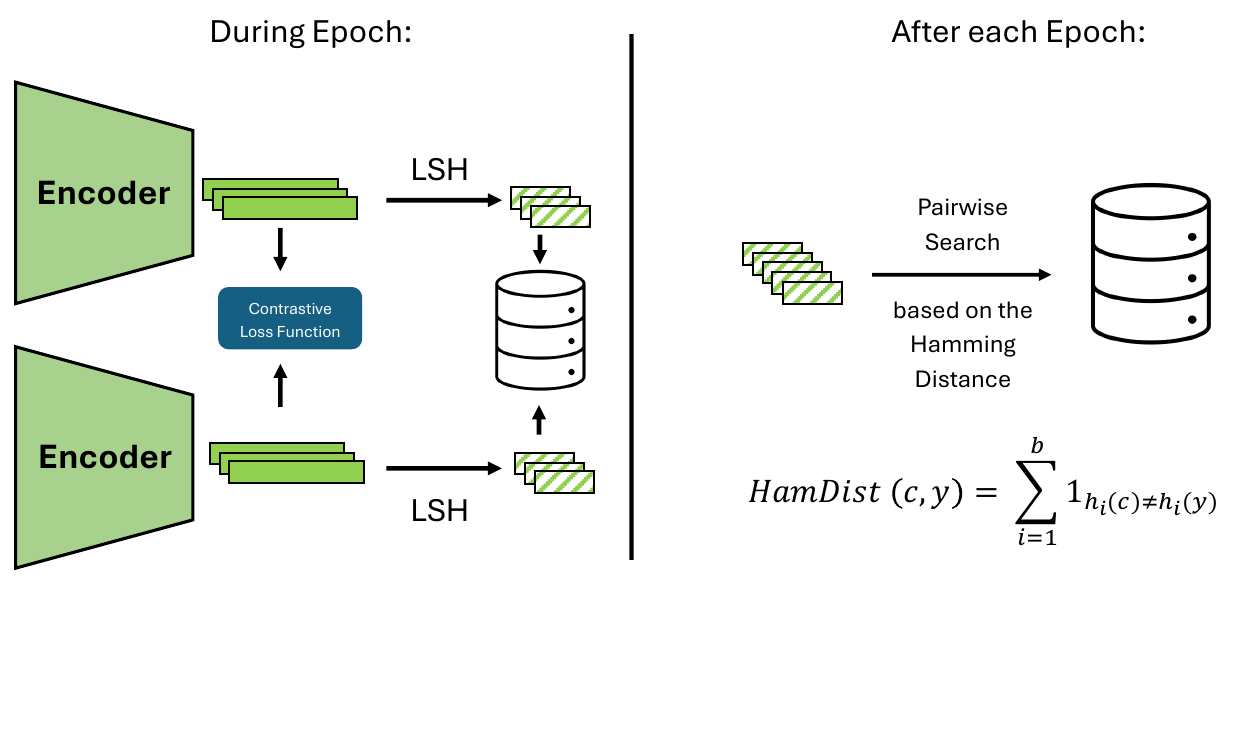}
    \caption{Framework for encoding input data and leveraging \ac{lsh} for binary transformation and \ac{hn} sampling during contrastive learning.}
    \label{fig:contrastive_lsh}
\end{figure}

\Cref{fig:contrastive_lsh} illustrates the process of encoding arbitrary input data, such as text or images, using an encoder to generate embeddings. \ac{lsh} is applied to transform these embeddings into binary vectors. After each epoch, pairwise search based on Hamming distance is used to sample \acp{hn}, for the next epoch. Similar to sampling on the float32 embedding, we define the \ac{hn} as the negative sample with the smallest distance to the anchor.
\subsection{LSH Design Choices}
\label{sec:lshdesignchoices}
\begin{wrapfigure}{r}{0.5\linewidth}
    \centering
    \includegraphics[width=\linewidth]{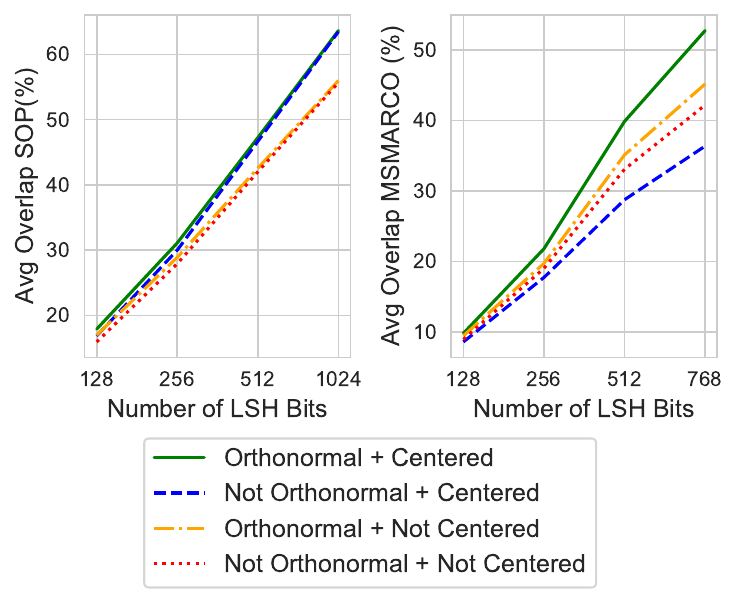}
    \caption{Impact of our \ac{lsh} design choices on the overlap for SOP and MS MARCO.}
    \label{fig:qr_mean_comp}
    \vspace{-3em}
\end{wrapfigure}

In \Cref{sec:overlap} we showed how a smaller bit size results in less overlap with the actual \ac{nn}, and as we can see in our performance evaluation, \Cref{tab:recall_comparison}, less overlap results in worse benchmark scores on the dataset. We now want to investigate how the design choices of our method improve this overlap. As described in \Cref{sec:lsh}, we use a random rotation matrix with orthonormal vectors and center the projected features. To obtain orthonormal vectors, we use QR decomposition. As shown in \Cref{fig:qr_mean_comp}, centering consistently improves the overlap between the found \ac{nn} and the actual \ac{nn}, especially in lower dimensional hash spaces, by reducing the skew in the binary representations.

For the MS MARCO dataset, the use of orthonormal matrices becomes increasingly important at higher bit dimensions, preserving feature variance and improving alignment with neighbours retrieved with the cosine similarity. Without orthonormalization, overlap performance degrades, especially in high-dimensional spaces. 

\subsection{Architecture Details:} 
\label{sec:archdetails}
For our experiments on image datasets, we use the ConvNeXt base model~\cite{liu2022convnet}, pre-trained on ImageNet-21k, from the timm library~\cite{rw2019timm}. ConvNeXt modernizes the ResNet architecture by incorporating design principles from the Vision Transformer~\cite{dosovitskiy2020image}. The model outputs 1024-dimensional embeddings and consists of 88 million parameters.

For our experiments on the MS MARCO text retrieval dataset, we use Distill-RoBERTa-base~\cite{Sanh2019DistilBERTAD}, a distilled version of RoBERTa~\cite{liu2019roberta}, with 82 million parameters. The hidden size of the transformer is 768. In both cases, we choose these models to allow efficient training and evaluation of our methodology. Additionally, we employ shared weights for both reference and query inputs.

\subsection{Training Times:} 
To quantify the computational efficiency of our proposed sampling strategy during training, we report the training time per epoch excluding the search time and the search time introduced by LSH-based sampling and pre-epoch \ac{hn} mining across different datasets. As shown in Table~\ref{tab:time_comparison}, the search time introduced by LSH-based sampling remains minimal across all datasets and represents a small fraction of the total cost per epoch. In contrast, pre-epoch \ac{hn} mining introduces significantly higher overheads, especially for large datasets such as MS MARCO, where search time dominates the training cost, accounting for over 90\% of the total. These results highlight the practical benefits of our approach during training. All search and training times in this comparison are averaged over 3 epochs.
\begin{table}[]
    \begin{center}
    \caption{Training and search time comparison across datasets. Times reported for 128-bit LSH and Pre-Epoch HN. Proportions are relative to training time per epoch.}
    \resizebox{\textwidth}{!}{
    \begin{tabular}{l|c|c|c|c|c|c} \hline \hline
        Dataset & \# Train Samples & Train Time (s) & LSH Time (s) & HN Time (s) & LSH (\%) & HN (\%) \\ \hline
        InShop & 25,882 & 65.57 & 0.21 & 13.33 & 0.32\% & 16.89\% \\
        CVUSA/CVACT & 35,532 & 181.91 & 0.35 & 25.10 & 0.19\% & 12.12\% \\
        VIGOR & 40,007 & 395.20 & 0.43 & 41.10 & 0.11\% & 9.42\% \\
        SOP & 59,551 & 137.24 & 0.71 & 69.04 & 0.52\% & 33.46\% \\
        MS MARCO & 532,736 & 566.92 & 33.57 & 7553.27 & 5.59\% & 93.01\% \\ \hline \hline
    \end{tabular}
    }
    \label{tab:time_comparison}
    \end{center}
\end{table}

\begin{figure}[]
    \centering
    \includegraphics[width=\linewidth]{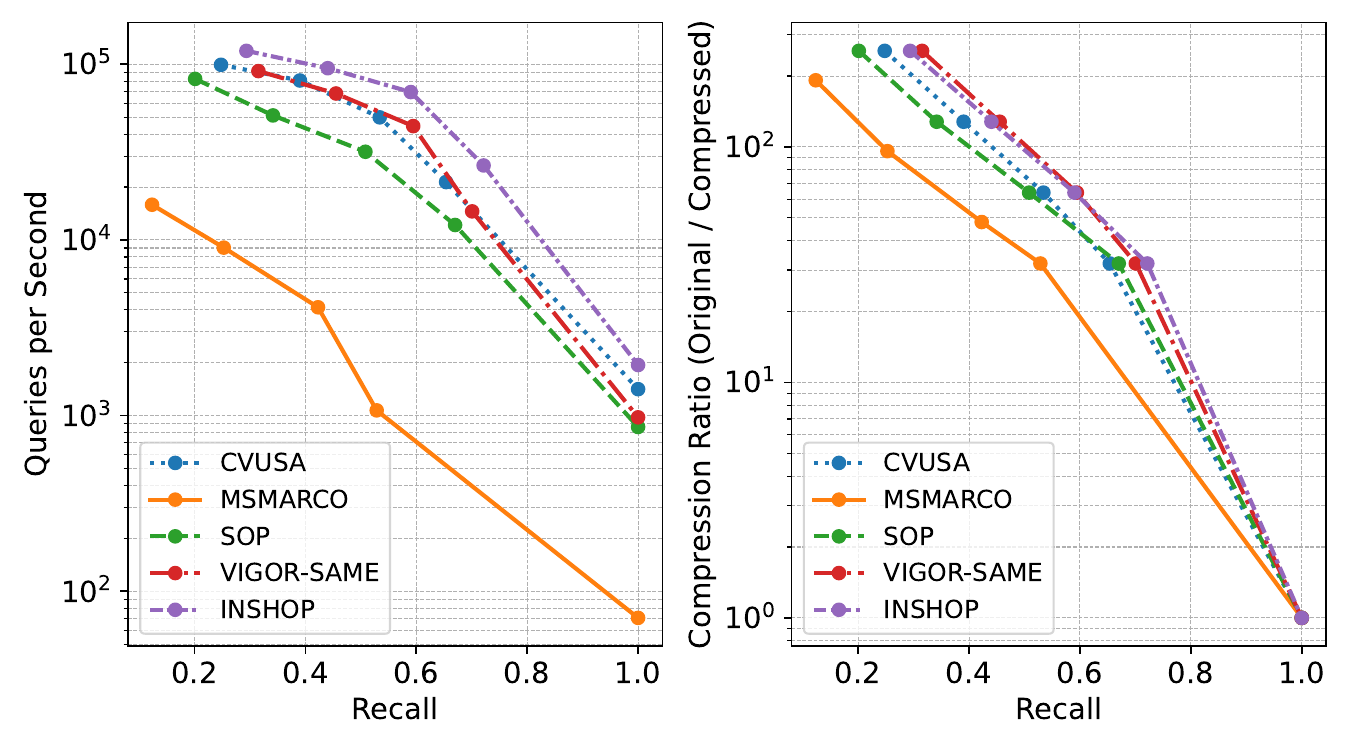}
    \caption{Comparison of true \ac{hn} neighbor recall with query throughput (left) and compression ratio (right) across five datasets. Curves reflect different LSH bit sizes (128, 256, 512, 768/1024), with vision datasets (CVUSA, SOP, VIGOR-SAME, INSHOP) using up to 1024 bits and the text dataset (MSMARCO) using up to 768 bits. Higher recall typically reduces query speed and compression efficiency. A recall of 1.0 indicates cosine similarity without LSH.}
    \label{fig:recallqps}
\end{figure}

In addition, we explore the tradeoffs in terms of queries per second and compression ratio versus the recall of \acp{hn} in \Cref{fig:recallqps}. On the left side, the general trend shows that increasing recall comes at the cost of decreasing query throughput, as higher bit rates or full-precision search methods (e.g. cosine similarity without \ac{lsh}) require more computation. On the right side, a similar trade-off is observed with the compression ratio, where higher recall requires storing more bits per vector, resulting in lower compression efficiency. Notably, the vision dataset allows for more aggressive compression while maintaining reasonable recall, while the text dataset (MSMARCO) shows a sharper performance degradation as recall increases. This effect is influenced by both dataset size, where larger datasets require proportionally larger indexes, leading to slower queries, and modality. In particular, identifying true \acp{hn} is more difficult for text datasets, where semantic similarities are more subtle compared to image datasets.
\subsection{PCA Comparison}
\label{sec:pcacomp}
\begin{wrapfigure}{r}{0.5\linewidth}
    \centering
    \includegraphics[width=\linewidth]{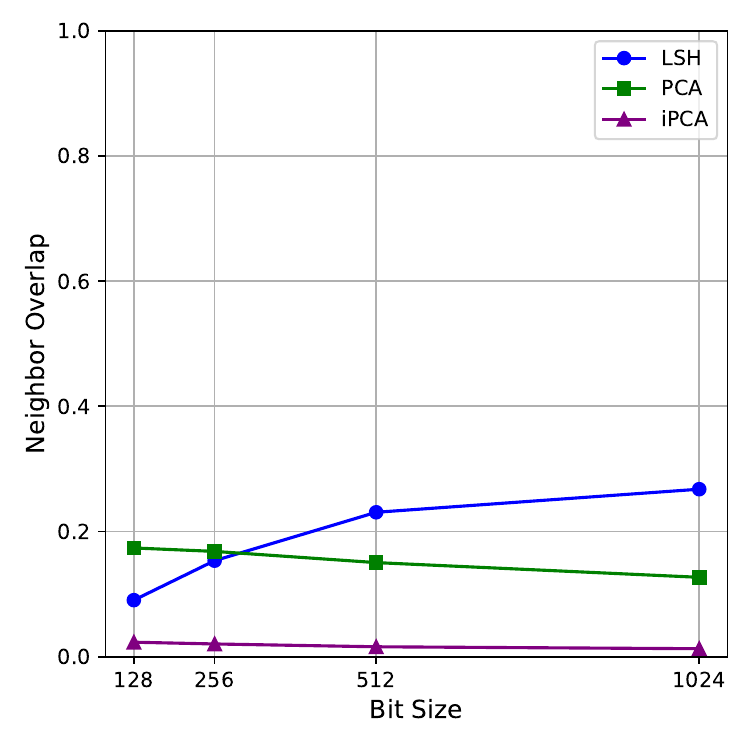}
    \caption{Comparison between \ac{lsh}, \ac{pca}, and incremental \ac{pca} in terms of overlap with the actual hardest neighbors.}
    \label{fig:pcavslsh}
    \vspace{-2em}
\end{wrapfigure}

We compare \ac{lsh} with a classical dimensionality reduction technique, namely \ac{pca}. For this experiment, shown in \Cref{fig:pcavslsh}, we used the CVUSA dataset and extracted features prior to any contrastive training using a ConvNeXt-base~\cite{liu2022convnet} model pre-trained on ImageNet. The extracted features were then projected to the target dimensionality, corresponding to the subspace size for \ac{pca} and the hash size for \ac{lsh}, and then binarized through a sign function to obtain compact binary codes. The comparison measures the overlap between the neighbors identified by \ac{pca} and \ac{lsh} and the true neighbors determined by the cosine similarity matrix.
As shown, \ac{pca} achieves slightly better performance at lower bit rates, but is surpassed by \ac{lsh} at higher bit rates. However, this comparison is somewhat unbalanced: unlike \ac{pca}, our \ac{lsh}-based approach is designed to operate during training without storing the original feature vectors, thereby significantly reducing memory consumption. For a more fair comparison, we also evaluated incremental \ac{pca}, where we store only 8 times batch size (128) samples to fit the PCA model before applying it to the remaining data. However, this setup results in a large performance penalty, as the limited sample size prevents PCA from capturing sufficient dataset statistics.

In addition to space requirements, \ac{pca} introduces additional training overhead, since fitting requires access to the entire dataset. This adds another computational burden, making it less suitable for our intended online training scenario. Therefore, we decided to exclude \ac{pca}-based approaches.

We did not include further experiments with data-dependent methods like Product Quantization~\cite{jegou2011pq} or Hierarchical Navigable Small World graphs~\cite{chong2019hsnw} as they would require constant re-training as well.

\end{document}